%% file: tsucb_wip.tex
\documentclass[11pt, letterpaper]{article}

\usepackage[final]{microtype}
\usepackage{color}
\usepackage[ruled]{algorithm2e}
\usepackage{varwidth}
\usepackage{capt-of}
\usepackage{boldline}
\usepackage{epsfig}
\usepackage{pgf}
\usepackage{latexsym}
\usepackage{url}
\usepackage{float} 
\usepackage[hypertexnames=false,hyperfootnotes=false]{hyperref}
\usepackage{texnansi}
\usepackage{color}
\usepackage{tikz}
\usepackage{afterpage}
\usepackage{enumerate}
\usepackage[normalem]{ulem}
\usepackage{lmodern}
\usepackage{rotating}
\usepackage[ruled]{algorithm2e}
\usepackage[noend]{algpseudocode}
\usepackage{graphicx}
\usepackage{caption} 
\usepackage{bbm}
\usepackage{layouts}
\usepackage{booktabs}

\usepackage{sectsty}
\usepackage{ifthen}

\usepackage[leqno]{amsmath}
\usepackage{amsthm}
\usepackage{amssymb}
\usepackage{amsfonts}
\usepackage{mathtools}
\usepackage[margin=10pt,font=small,labelfont=bf]{caption}

\usepackage{enumitem}
\usepackage{dsfont}
\usepackage{subfigure}
\usepackage{bm}
\usepackage{xparse} 

\usepackage{natbib}
 \bibpunct[, ]{(}{)}{,}{a}{}{,}%

\makeatletter
\g@addto@macro{\UrlBreaks}{\UrlOrds}
\makeatother

\DeclareMathOperator*{\argmin}{\mathrm{argmin}}
\DeclareMathOperator*{\argmax}{\mathrm{argmax}}

\newtheoremstyle{thm-sf}{}{}{\itshape}{}{\sffamily\bfseries}{.}{ }{}
\theoremstyle{thm-sf}

\newtheorem{example}{Example}
\newtheorem{theorem}{Theorem}

\newtheorem{corollary}{Corollary}
\newtheorem{lemma}{Lemma}

\newtheorem{proposition}{Proposition}

\input{commands}
\newcommand{\TableSpaced}{}

\NewDocumentEnvironment{myproof}{o}
  {\IfNoValueTF{#1}{\paragraph{{\normalfont {\normalsize \textit{Proof.}}}}} {\paragraph{{\small #1} }} }
  {\hfill$\square$}

\usepackage{xcolor}

\usetikzlibrary{arrows,patterns,plotmarks}
\tikzstyle{every picture} += [>=stealth]
\allsectionsfont{\sffamily}

\makeatletter
\def\@seccntformat#1{\csname the#1\endcsname.\quad}
\makeatother

\floatstyle{ruled}
\provideboolean{fastcompile}
\newcommand{\hidefastcompile}[1]{\ifthenelse{\boolean{fastcompile}}{}{#1}}

\usepackage[margin=1in]{geometry}
\usepackage{setspace}
\newcommand{\edit}[1]{#1}

\newtheorem*{le:ell}{Lemma~\ref{le:ell}}
\newtheorem*{le:feas}{Lemma~\ref{le:feas}}
\newtheorem*{le:sample_complexity}{Lemma~\ref{le:sample_complexity}}

\usepackage{setspace}

\title{\textsf{\textbf{TS-UCB: Improving on Thompson Sampling With Little to No Additional Computation}}}
\author{
Jackie Baek \\ Operations Research Center \\ Massachusetts Institute of Technology \\ \url{baek@mit.edu}
\and Vivek F. Farias \\ Sloan School of Management \\ Massachusetts Institute of Technology \\ \url{vivekf@mit.edu}
}
\date{}

\begin{document}
\maketitle






\begin{abstract}
\input{sections/abstract.tex}
\vskip 5pt
\end{abstract}

\setstretch{1.5}


\section{Introduction}
\input{sections/introduction.tex}

\section{Model} \label{sec:model}
\input{sections/model.tex}

\section{Algorithm} \label{sec:alg}
\input{sections/algorithm.tex}

\section{Computational Results} \label{sec:comp}
\input{sections/comp_results.tex}



\bibliographystyle{informs2014} 
\bibliography{../ref} 




\newpage
\appendix
\section{Regret Analysis} \label{app:A}
\input{sections/regret_analysis.tex}

\end{document}

%% file: commands.tex
\newcommand{\bI}{\mathds{1}}

\newcommand{\bR}{\mathbb{R}}
\newcommand{\bE}{\mathbb{E}}

\newcommand{\ssa}{a^*}

\newcommand{\eps}{\epsilon}

\newcommand{\ii}{\textit}
\newcommand{\bb}{\textbf}

\newcommand{\ttheta}{\tilde{\theta}}

\newcommand{\rad}{\mathrm{radius}}
\newcommand{\sumt}{\sum_{t=1}^T}
\newcommand{\sumtK}{\sum_{t=K+1}^T}
\newcommand{\BReg}{\mathrm{BayesRegret}}
\newcommand{\Reg}{\mathrm{Regret}}

\newcommand{\ssA}{A^*}

\newcommand{\hmu}{\hat{\mu}}
\newcommand{\cA}{\mathcal{A}}

\newcommand{\ft}{f_{\theta}}
\newcommand{\tf}{\tilde{f}}

\newcommand{\cH}{\mathcal{H}}
\newcommand{\bX}{\mathbf{X}}
\newcommand{\bY}{\mathbf{Y}}
\newcommand{\htt}{\hat{\theta}}

\newcommand{\Psie}{\Psi}
\newcommand{\bPsi}{\bar{\Psi}}
\newcommand{\rmin}{r_{\min}}
\newcommand{\rmax}{r_{\max}}
\newcommand{\subg}{r}

\newcommand{\var}{\text{Var}}
\newcommand{\TSalg}{\textsc{TS}}
\newcommand{\TSUCB}{\textsc{TS-UCB}}
\newcommand{\IDS}{\textsc{IDS}}

\newcommand{\TSUCBE}{\textsc{TS-UCB}}
\newcommand{\At}{A^{\TSUCB}_t}

%% file: sections/abstract.tex
Thompson sampling has become a ubiquitous approach to online decision problems with bandit feedback. The key algorithmic task for Thompson sampling is drawing a sample from the posterior of the optimal action.
We propose an alternative arm selection rule we dub $\TSUCB$, that requires negligible additional computational effort but provides significant performance improvements relative to Thompson sampling.
At each step, $\TSUCB$ computes a score for each arm using two ingredients: posterior sample(s) and upper confidence bounds.
$\TSUCB$ can be used in any setting where these two quantities are available, and it is flexible in the number of posterior samples it takes as input.
$\TSUCB$ achieves materially lower regret on a comprehensive suite of synthetic and real-world datasets, 
including a personalized article recommendation dataset from Yahoo! and a suite of benchmark datasets from a deep bandit suite proposed in \cite{riquelme2018deep}.
Finally, from a theoretical perspective, we establish optimal regret guarantees for $\TSUCB$ for both the $K$-armed and linear bandit models.

%% file: sections/introduction.tex
This paper studies the stochastic multi-armed bandit problem, a classical problem modeling sequential decision-making under uncertainty.
This problem captures the inherent tradeoff between exploration and exploitation.
We study the Bayesian setting, in which we are endowed with an initial prior on the mean reward for each arm.


Thompson sampling (TS) \citep{thompson1933likelihood}, has in recent years come to be a solution of choice for the multi-armed bandit problem. This popularity stems from the fact that the algorithm performs well empirically \citep{scott2010modern,chapelle2011empirical} and also admits near-optimal theoretical performance guarantees \citep{agrawal2012analysis,agrawal2013thompson,kaufmann2012thompson,bubeck2013prior,russo2014learning,russo2016information}. Perhaps one of the most attractive features of Thompson sampling though, is the simplicity of the algorithm itself: the key algorithmic task of TS is to sample once from the posterior on arm means, a task that is arguably the simplest thing one can hope to do in a Bayesian formulation of the multi-armed bandit problem.

\textbf{This Paper: }Against the backdrop of Thompson sampling, we propose $\TSUCB$. Given one or more samples from the posterior on arm means, $\TSUCB$ simply provides a distinct approach to scoring the possible arms.
The only additional ingredient this scoring rule relies on is the availability of so-called upper confidence bounds (UCBs) on these arm means.

Now both sampling from a posterior, as well as computing a UCB can be a potentially hard task, especially in the context of bandit models where the payoff from an arm is a complex function of unknown parameters. A canonical example of such a hard problem variant is the contextual bandit problem wherein mean arm reward is given by a complicated function (say, a deep neural network) of the context. \cite{riquelme2018deep} provide a recent benchmark comparison of ten different approaches to sampling from an approximate posterior on unknown arm parameters. They show that an approach that chooses to model the uncertainty in \ii{only the last layer} of the neural network defining the mean reward from pulling a given arm at a given context is an effective and robust approach to posterior approximation. In such an approach, not only is (approximate) posterior sampling possible, but UCBs have a closed-form expression and can be easily computed, making possible the use of $\TSUCB$.

\textbf{Our Contributions: }We show that $\TSUCB$ provides material improvements over Thompson sampling {\em across the board} on comprehensive sets of synthetic and real-world datasets.
The real-world datasets include personalizing news article recommendations for the front page of Yahoo!, and a benchmark set of deep bandit problems studied in \cite{riquelme2018deep}.
Importantly, the performance of $\TSUCB$ either matched or improved upon the state-of-the-art algorithm Information-Directed Sampling (IDS) \citep{russo2018ids}, which requires approximately three orders of magnitude more sampling (and thus compute) than either TS or $\TSUCB$.

$\TSUCB$'s arm scoring rule can be seen as a modification of the one used in IDS.
In particular, $\TSUCB$ essentially replaces the role of the ``information gain'' term used in IDS to a quantity that is much easier to compute: the radius of the confidence interval.
This modification makes $\TSUCB$ orders of magnitude cheaper in terms of computation than IDS, and the experimental results show that this does not come at the cost of any degradation in performance.
Another interpretation of $\TSUCB$ is that it is a UCB algorithm, but one that \ii{automatically} and \ii{dynamically} tunes the parameter that controls the width of the confidence interval.
This is the first algorithm of such kind to the best of our knowledge.

Theoretically, we analyze $\TSUCB$ in two specific bandit settings: the $K$-armed bandit and the linear bandit.
In the first setting, there are $K$ independent arms.
In the linear bandit, each arm is a vector in $\bR^d$, and the rewards are linear in the chosen arm.
In both settings, $\TSUCB$ is agnostic to the time horizon.
We prove the following Bayes regret bounds for $\TSUCB$:

{\it For the $K$-armed bandit, the Bayes regret of $\TSUCB$ is at most $O(\sqrt{KT\log T})$.}

{\it For the linear bandit of dimension $d$, the Bayes regret of $\TSUCB$ is at most $O(d\log T \sqrt{T})$.}

Both of these results match the lower bounds up to log factors.
The results are stated formally in Theorems~\ref{thm:mainK} and \ref{thm:mainlinear}.

\subsection{Related Literature}
Given the vast literature on bandit algorithms, we restrict our review to literature heavily related to our work, viz. literature focused on the development and analysis of UCB algorithms, literature analyzing Thompson sampling (TS), literature on linear contextual bandits, and literature on methods of applying deep learning models to bandit problems.

The UCB algorithm \citep{auer2002finite} computes an upper confidence bound for every action, and plays the action whose UCB is the highest.
In the Bayesian setting, `Bayes UCB' is defined as the $\alpha$'th percentile of this distribution, and \cite{kaufmann2012bayesian} show that using $\alpha = 1-\frac{1}{t \log^c t}$ achieves the lower bound of \cite{lai1985asymptotically} for $K$-armed bandits.
For linear bandits, \cite{dani2008stochastic} prove a lower bound of $\Omega(d \sqrt{T})$ for infinite action sets, and the UCB algorithms from \cite{dani2008stochastic,rusmevichientong2010linearly,abbasi2011improved} match this up to log factors.  It is worth noting that neither the UCB or Bayes UCB algorithms are competitive on the benchmark set of problems in \cite{riquelme2018deep}.

As discussed, TS is a randomized Bayesian algorithm that chooses an action with the same probability that the action is optimal.
Though it was initially proposed in \cite{thompson1933likelihood}, TS has only recently gained a surge of interest,
largely influenced by the strong empirical performance of TS demonstrated in \cite{chapelle2011empirical} and \cite{scott2010modern}.
Since then, many theoretical results on regret bounds for TS have been established \citep{agrawal2012analysis,agrawal2013further,agrawal2013thompson,agrawal2017near,kaufmann2012thompson}.
In the Bayesian setting, \cite{russo2014learning} prove a regret bound of $O(\sqrt{KT\log T})$ and $O(d \log T \sqrt{T})$ for TS in the $K$-armed and linear bandit setting respectively. \cite{bubeck2013prior} improve the regret in the Bayesian $K$-armed setting to $O(\sqrt{KT})$, and they show this is order-optimal. 

The ideas in this paper were heavily influenced by our reading of \cite{russo2014learning, russo2018ids}. In the former paper, the authors use UCB algorithms as an {\em analytical} tool to analyze TS. This begs the natural question of whether an appropriate decomposition of regret can provide insight on {\em algorithmic} modifications that might improve upon TS. \cite{russo2018ids} provide such a decomposition and proposes Information Directed Sampling (IDS). IDS has been shown to provide significant performance improvement over TS in some cases, but has heavy sampling (and thus, computational) requirements. The present paper presents yet another decomposition, providing an arm selection rule that does not require additional sampling (i.e. a single sample from the posterior continues to suffice), but nonetheless provides significant improvements over TS while being competitive with IDS.


\edit{
Our simulations consider the contextual bandit problem, a setting with a wide variety of applications including healthcare \citep{bastani2020online}, recommendation systems \citep{agarwal2009online}, and dynamic pricing \citep{ban2020personalized}.
Specifically, we consider linear contextual bandits, which can be viewed as a special case of linear bandits, and therefore the same $O(d \log T \sqrt{T})$ regret bound for TS can be applied.
However, it is known that by making additional assumptions on the context generation process, one can achieve logarithmic regret bounds with $\eps$-greedy \citep{goldenshluger2013linear,bastani2020online} or greedy algorithms \citep{bastani2020mostly,kannan2018smoothed}.
These algorithms depend on the randomness of the contexts to provide the exploration needed to learn the unknown parameters.
We run the greedy algorithm as one of the benchmark policies in our simulations.
}

On the deep learning front, 
one key idea that has been used to apply deep learning to sequential decision making problems
is to use TS \citep{riquelme2018deep,lu2017ensemble,dwaracherla2020hyper}.
Since TS requires just a single sample from the posterior, if the posterior can be approximated in some way, then TS can be readily applied.
\cite{riquelme2018deep} use this idea and evaluates TS on ten different posterior approximation methods for neural networks, ranging from variational methods \citep{graves2011practical}, MCMC methods \citep{neal2012bayesian}, among others.
The authors find that the approach of modeling uncertainty on just the last layer of the neural network (the `Neural-Linear' approach) \citep{snoek2015scalable,hinton2008using,calandra2016manifold} was overall one of the most effective approaches. This neural linear approach provides not just a tractable approach to approximate posterior sampling, but further provides a tractable UCB for the problem as well. As such, the neural linear approach facilitates the use of the TS-UCB arm selection rule, and we show that TS-UCB provides significant improvements over the use of TS on the deep bandit benchmark in \cite{riquelme2018deep}.


The rest of the paper is structured as follows.
We describe the bandit model in Section~\ref{sec:model}, and we define the $\TSUCB$ policy in Section~\ref{sec:alg}, where we also state the main theoretical results.
In Section~\ref{sec:comp}, we show the results of computational experiments on sythetic and real-world datasets.
We give an outline of the regret analysis in Section~\ref{sec:pfsketch}, and the formal proofs can be found in Appendix~\ref{app:A}.

%% file: sections/model.tex
Let $\cA$ be a compact set of all possible actions.
At time $t$, an agent is presented with a possibly random subset $\cA_t \subseteq \cA$ in which they choose an action to play from.
If action $a$ is chosen at time $t$, the agent immediately observes a random reward $R_t(a) \in \bR$.
For each action $a$, the sequence $(R_t(a))_{t \geq 1}$ is i.i.d. and independent of plays of other actions.
The mean reward of each action $a$ is $f_\theta(a)$, where $\theta \in \Theta$ is an unknown parameter, and $\{f_\theta:\cA \rightarrow \bR | \theta \in \Theta\}$ is a known set of deterministic functions.
That is, $\bE[R_t(a)|\theta] = f_\theta(a)$ for all $a \in \cA$ and $t \geq 1$.

Let $H_t = (\cA_1, A_1, R_1(A_1), \dots, \cA_{t-1}, A_{t-1}, R_{t-1}(A_{t-1}), \cA_t)$ denote the history of observations available when the agent is choosing the action for time $t$, and let $\cH$ denote the set of all possible histories.
We often refer to $H_t$ as the ``state'' at time $t$.
A policy $(\pi_t)_{t \geq 1}$ is a deterministic sequence of functions mapping the history to a distribution over actions.
An agent employing the policy plays the random action $A_t$ distributed according to $\pi_t(H_t)$, where $H_t$ is the current history.
We will often write $\pi_t(a)$ instead of $\pi_t(H_t)(a)$, where $\pi_t(a) = \Pr(A_t = a | H_t)$.
Let $\ssA_t:\Theta \rightarrow \cA_t$ be a function satisfying $\ssA_t(\theta) \in \argmax_{a \in \cA_t} f_\theta(a)$, which represents the optimal action at time $t$ if $\theta$ were known.
We use $\ssA_t$ to denote the random variable $\ssA_t(\theta)$, where $\theta$ is the true parameter.

The $T$-period regret of policy $\pi$ is defined as
\[
\Reg(T, \pi, \theta) = \sumt \bE[f_\theta(\ssA_t) - f_\theta(A_t) | \theta].
\]
We study the Bayesian setting, in which we are endowed with a known prior $q$ on the parameter $\theta$.
We take an expectation over this prior to define the $T$-period Bayes regret
\begin{align*}
\BReg(T, \pi) &= \sumt \bE[f_\theta(\ssA_t) - f_\theta(A_t)].
\end{align*}
We assume that the agent can perform a Bayesian update to their prior at each step after the reward is observed.
Let $q(H_t)$ denote to the posterior distribution of $\theta$ given the history $H_t$.
In our work, we assume that the agent is able to \ii{sample} from the distribution $q(H_t)$ for any state $H_t$.

We end this section by describing two concrete bandit models that are the focus of our regret analysis.

\subsection{K-armed Bandit}
In this setting, $\cA_t = \cA = [K]$ for all $t$, and each of the entries of the unknown parameter $\theta \in \bR^K$ correspond to the mean of each action.
That is, $\ft(i) = \theta_i$ for every $i \in [K]$.
We assume that $\theta_a \in [0, 1]$ for all $a$, and the rewards $R_t(a)$ are also bounded in $[0, 1]$ for all $a$ and $t$.
The prior distribution $q$ on $\theta$, supported on $[0, 1]^K$, can otherwise be arbitrary.

\subsection{Linear Bandit}
In the linear bandit setting, there is a known vector $X(a) \in \bR^d$ associated with each $a \in \cA$, and the mean reward takes on the form $\ft(a) = \langle \theta, X(a) \rangle$, for $\theta \in \Theta \subseteq \bR^d$.
We assume that $||\theta||_2 \leq S \leq \sqrt{d}$, $||X(a)|| \leq L$, and $\ft(a) \in [-1, 1]$ for all $a \in \cA$.
Lastly, we assume that $R_{t}(a) - \ft(a)$ is $\subg$-sub-Gaussian for every $t$ and $a$ for some $\subg \geq 1$.
All of these assumptions are standard and are the same as in \cite{abbasi2011improved}.

A special case of linear bandits is contextual linear bandits in which there are $K$ arms and there is an unknown parameter $\beta_k \in \bR^d$ for each arm $k \in [K]$.
A random context $X_t \in \bR^d$ is observed at the start of each time step, and the mean reward for arm $k$ at time $t$ is $\langle \beta_k, X_t \rangle$.
This is equivalent to a linear bandit problem of dimension $dk$ where the action set (transposed) at time $t$ is
$\{ (X_t^\top, 0_d, \dots ,0_d), (0_d, X_t^\top, \dots, 0_d), \dots, (0_d, \dots, 0_d, X_t^\top)\}$, where $0_d$ is the transposed 0-vector of dimension $d$, and the unknown parameter is $\theta^{\top} = (\beta_1^{\top}, \dots, \beta_K^{\top})$ of dimension $dk$.


%% file: sections/algorithm.tex
$\TSUCB$ requires a set of functions $U,\hmu: \cH \times \cA \rightarrow \bR$ to first be specified, where $U(h, a)$ represents the upper confidence bound of action $a$ at history $h$, and $\hmu(h, a)$ represents an estimate of $f_\theta(a)$ at history $h$.
We require that $U(h, a) - \hmu(h, a) > 0$ on every input.
We write $U_t(a) = U(H_t, a)$ and $\hmu_t(a) = \hmu(H_t, a)$,
and we refer to the quantity $\rad_t(a) \triangleq U_t(a) - \hmu_t(a)$ as the \ii{radius} of the confidence interval.


$\TSUCB$ proceeds as follows.
At state $H_t$, draw $m$ independent samples from the posterior distribution $q(H_t)$, for some integer parameter $m \geq 1$.
Denote these samples by $\ttheta_1, \dots, \ttheta_m$, and let $\tf_i = f_{\ttheta_i}(\ssA_t(\ttheta_i))$
be the mean reward of the best arm when the true parameter is $\ttheta_i$.
(Conditioned on $H_t$, the distribution of $\tf_i$ is the same as the distribution of $\ft(\ssA)$.)
Let $\tf_t = \frac{1}{m} \sum_{i = 1}^m \tf_i$.
For every action $a$, define the ratio $\Psi_t(a)$ as
\begin{align} \label{eq:defratio1}
\Psi_t(a) \triangleq \frac{\tf_t - \hmu_{t}(a) }{U_t(a) - \hmu_{t}(a)}
= \frac{\tf_t - \hmu_{t}(a) }{\rad_{t}(a)}.
\end{align}
$\TSUCB$ chooses an action that minimizes this ratio, which we assume exists.\footnote{Clearly it exists if $\cA$ is finite. Otherwise, since $\cA$ is assumed to be compact, it exists if $\hmu_t$ and $U_t$ are continuous functions.}
That is, if $A^{\TSUCB}_t$ is the random variable for the action chosen by $\TSUCB$ at time $t$, then,
\begin{align}
A^{\TSUCB}_t \in \argmin_{a \in \cA_t} \Psi_t(a).
\end{align}
We parse the ratio $\Psi_t(a)$:
$\hmu_{t}(a)$ is an estimate of the expected reward $\bE[\ft(a)|H_t]$ from playing action $a$, and $\tf_t$ is an estimate of the optimal reward $\bE[\ft(\ssA)|H_t]$ (indeed, $\tf_t \rightarrow \bE[\ft(\ssA)|H_t]$ as $m \rightarrow \infty$).
Then, the numerator of the ratio estimates the expected instantaneous regret from playing action $a$.
We clearly want this to be small, but minimizing only the numerator would result in the greedy policy. 
The denominator enforces exploration by favoring actions with larger confidence intervals, corresponding to actions in which not much information is known about.
This ratio is similar to the information ratio that is minimized in the IDS algorithm --- the main difference is that the denominator in IDS is information gain.

$\TSUCB$ can be applied whenever the quantities $\tf_t = \frac{1}{m} \sum_{i = 1}^m \tf_i$ and $\{U_t(a), \hmu_t(a)\}_{a \in \cA}$ can be computed,
which are exactly the quantities needed for TS ($m=1$) and UCB respectively.
The following example shows that $\TSUCB$ can be applied in a general setting where the relationship between actions and rewards is modeled using a deep neural network.
\begin{example}[Neural Linear \citep{riquelme2018deep}] \label{ex:neural_linear}
Consider a contextual bandit problem where a context $X_t \in \bR^{d'}$ arrives at each time step, and the expected reward of taking action $a \in \cA$ is $g(X_t, a)$, for an unknown function $g$.
The `Neural Linear' method models uncertainty in only the last layer of the network by considering a specific class of functions $g$.
Specifically, consider that $g$ allows the decomposition $g(X_t, a) = h(X_t)^\top \beta_a$ where $h(X_t) \in \bR^{d}$ represent the outputs from the last layer of some neural network and $\beta_a \in \bR^d$ is some parameter vector. If the function $h(\cdot)$ were known, then the resulting problem is a linear bandit problem for which both sampling from the posterior on $\beta_a$ for all $a \in \cA$ as well as computing a (closed form) UCB on $\beta_a$ are easy.
In reality $h(\cdot)$ is unknown but the Neural Linear method approximates this quantity from past observations and ignores uncertainty in the estimate. As such, it is clear that TS-UCB can be used as an alternative to TS in the Neural Linear approach.
\end{example}

We evaluate the method described in the above example on a range of real-world datasets in Section~\ref{sec:exp_real_world}.

\subsection{Interpretation as a Dynamic UCB Algorithm}
Recall that a UCB algorithm chooses the arm with the highest upper confidence bound.
Often, the radius of the confidence interval takes the form $\alpha \cdot \rad_t(a)$, where $\alpha$ is a scalar parameter.
For example, the UCB1 algorithm of \cite{auer2002finite} plays the arm that maximizes $\hmu_t(a) + \sqrt{\frac{2 \log t}{N_t(a)}}$; here we can think of $\alpha = \sqrt{2}$.
It is well known that tuning this parameter can vastly improve empirical performance \citep{russo2014learning}.

$\TSUCB$ can be interpreted as a UCB algorithm whose $\alpha$ parameter is dynamically tuned.
$\TSUCB$ plays the arm with the highest $\hmu_t(a) + \alpha_t \cdot \rad_t(a)$, where
\begin{align} \label{eq:alpha_t}
\alpha_t = \min \{\alpha : \max_{a \in \cA}  \{\hmu_t(a) + \alpha \cdot \rad_t(a) \} \geq \tf_t \}.
\end{align}
That is, after sampling $\tf_t$,
$\alpha_t$ is the smallest $\alpha$ such that there exists an arm whose UCB, $\hmu_t(a) + \alpha \cdot \rad_t(a)$, is at least as large as $\tf_t$.
This ends up being equivalent to setting $\alpha_t = \Psi_t(A^{\TSUCB}_t)$.
Indeed, for any $a \in \cA_t$, since $\Psi_t(A^{\TSUCB}_t) \leq \Psi_t(a)$ by definition of the algorithm, we have
\begin{align*}
\hmu_t(a) + \Psi_t(A^{\TSUCB}_t) \cdot \rad_t(a)
\leq \hmu_t(a) + \Psi_t(a) \cdot \rad_t(a)
= \tf_t,
\end{align*}
where the inequality is an equality if and only if $a \in \argmin_{a \in \cA} \Psi_t(a)$.
In other words, for the action that $\TSUCB$ chooses, its (dynamically tuned) UCB is exactly $\tf_t$; for other actions, their UCB is smaller.
In this sense, $\TSUCB$ is a method of \ii{automatically} (since the parameter $\alpha_t$ is adjusted using posterior samples) and \ii{dynamically} (since $\alpha_t$ changes with $t$) tuning the UCB algorithm.

We now apply $\TSUCB$ for the $K$-armed bandit and linear bandit using the standard definitions of upper confidence bounds found in the literature, and we formally state the main theorems.

\subsection{K-armed Bandit}
We assume $T \geq K$, and we pull every arm once in the first $K$ time steps.
Let $N_t(a) = \sum_{s = 1}^{t-1} \bI(A_s = a)$ be the number of times that action $a$ was played up to but not including time $t$.
We define the upper confidence bounds in a similar way to \cite{auer2002finite}; namely,
\begin{align}
  \hmu_t(a) &\triangleq \frac{1}{N_t(a)} \sum_{s = 1}^{t-1} \bI(A_s = a) R_s(a)
  & U_t(a) &\triangleq \hmu_{t}(a) + \sqrt{\frac{3\log T}{N_{t}(a)}}.\label{eq:karmucb}
\end{align}
This implies $\rad_t(a) = \sqrt{\frac{3\log T}{N_{t}(a)}}$.

Because the term $\sqrt{3 \log T}$ appears as a multiplicative factor in the radius and the same term is used for all actions and time steps,
the algorithm is agnostic to this value.
That is, $\TSUCB$ reduces to picking the action which minimizes
\begin{align} \label{eq:karmsimpleform}
\sqrt{N_t(a)}(\tf_t - \hmu_{t}(a)).
\end{align}
This implies that $\TSUCB$ does not have to know the time horizon $T$ a priori.
We now state our main result for this setting.
\begin{theorem} \label{thm:mainK}
For the $K$-armed bandit, using the UCBs as defined in \eqref{eq:karmucb},
\begin{align}
 \BReg(T, \pi^{\TSUCBE}) &\leq  4 \sqrt{3 KT \log T} + T^{-2} + 3\sqrt{T} + K  =  O(\sqrt{ KT \log T}). \label{eq:mainke}
\end{align}
\end{theorem}
This result matches the $\Omega(\sqrt{KT})$ lower bound \cite{bubeck2013prior} up to a logarithmic factor.
It is worth noting that TS has been shown to match the lower bound exactly \cite{bubeck2013prior}; we believe that the logarithmic gap is a shortcoming of our analysis.

\subsection{Linear Bandit} \label{sec:linear_bandit}
For the linear bandit, to define the functions $\hmu_t$ and $U_t$, we first need to define a confidence set $C_t \subseteq \Theta$, which contains $\theta$ with high probability.
We use the confidence sets developed in \cite{abbasi2011improved}.
Let $X_t = X(A_t)$ be the vector associated with the action played at time $t$.
Let $\bX_t$ be the $t \times d$ matrix whose $s$'th row is $X_s^\top$.
Let $\bY_t \in \bR^t$ be the vector of rewards seen up to and including time $t$.
At time $t$, define the positive semi-definite matrix $V_t =  I + \sum_{s = 1}^t X_s X_s^\top = I + \bX_t^\top \bX_t$, and construct the estimate $\htt_t = V_t^{-1} \bX_t^\top \bY_t$.
Using the notation $||x||_A = \sqrt{x^\top A x}$, let $C_t = \{\rho : ||\rho - \htt_t||_{V_t} \leq \sqrt{\beta_t}\}$,
where $\sqrt{\beta_t} = \subg \sqrt{d \log(T^2(1+tL))} + S$.

Using this confidence set, the functions needed for $\TSUCB$ are defined as
\begin{align}
\hmu_t(a) &\triangleq \langle X(a), \htt_t \rangle &
U_t(a) &\triangleq \max_{\rho \in C_t} \langle X(a), \rho \rangle. \label{eq:linearucb}
\end{align}
Since $U_t(a)$ is the solution to maximizing a linear function subject to an ellipsoidal constraint, it has a closed form solution: $U_t(a) = \langle X(a), \htt_t \rangle + \sqrt{\beta_t} || X(a)||_{V_t^{-1}} $, which implies $\rad_t(a) = \sqrt{\beta_t} || X(a)||_{V_t^{-1}}$.
Then, $\TSUCB$ reduces to picking the action which minimizes
\begin{align*}
\frac{\tf_t - \langle X(a), \htt_t \rangle}{|| X(a)||_{V_t^{-1}}}.
\end{align*}
Note that the $\sqrt{\beta_t}$ term disappears, implying $\TSUCB$ does not depend on the exact expression of this term.
Like the $K$-armed bandit, the algorithm does not have to know the time horizon $T$ a priori.


We state our main result for this setting.

\begin{theorem} \label{thm:mainlinear}
For the linear bandit, using the UCBs as defined in \eqref{eq:linearucb},
if $||X(a)||_2 = 1$ for all $a \in \cA$,
\begin{align} \label{eq:mainlineare}
 \BReg(T, \pi^{\TSUCBE}) &\leq B+ T^{-2} + 12 \sqrt{2T}
             = O(d \log T \sqrt{T}).
\end{align}
where
\begin{align*}
 B =8\sqrt{Td \log(1+TL/d)}(S + \subg \sqrt{6 \log(T) + d \log(1 + T/d)})
 = O(d \log T \sqrt{T}).
\end{align*}
\end{theorem}
This result matches the $\Omega(d\sqrt{T})$ lower bound \citep{dani2008stochastic} up to a logarithmic factor.
We believe the additional assumption that $||X(a)||_2 = 1$ is an artifact our proof, which we believe can be likely removed with a more refined analysis. We note that TS and IDS has been shown to achieve a regret of $O(\sqrt{dT\log(|\cA|)})$ \citep{russo2016information,russo2018ids}, which is dependent on the total number of actions $|\cA|$.

We give an outline of the proofs of Theorem~\ref{thm:mainK} and \ref{thm:mainlinear} in Section~\ref{sec:pfsketch}, and we provide the full proof in Appendix~\ref{app:A}.

%% file: sections/comp_results.tex
We conduct three sets of experiments for the contextual bandit problem. The first set is entirely synthetic for an ensemble of linear contextual bandit problems where exact posterior samples (and a regret analysis) are available for all methods considered.
Our objective here is to understand the level of improvement $\TSUCB$ can provide over TS and how the level of this improvement depends on
natural problem features such as the number of arms and the level of noise.
The next two experiments are on real-world datasets, where the exact Bayesian structure is not available.
The second set of experiments considers the problem of personalizing news article recommendations on the front page of the Yahoo! website, and the last set of experiments considers a deep bandit benchmark on seven different real-world datasets.
Our goal is to show that $\TSUCB$ provides state of the art performance while being computationally cheap and robust to prior misspecification.
In all three experiments, we compare $\TSUCB$ to TS, UCB, Greedy, and IDS.

\subsection{Synthetic Experiments}
First, we simulate synthetic instances of the linear contextual bandit with varying number of actions and size of the prior covariance.
Let $d$ be the dimension and $K$ be the number of actions.
For each action $k \in [K]$, we independently sample $\beta_k \sim N(0, I_d)$, where $I_d$ is the $d$-dimensional identity matrix.
At each time $t$, a context $X_t$ is drawn i.i.d. from $N(0, \frac{1}{d} I_d)$.
The reward for arm $k$ at time $t$ is $\langle \beta_k, X_t \rangle +
\eps_t$, where $\eps_t$ is drawn i.i.d. from $N(0, \sigma^2)$.
We set $d = 10$ and vary the number of actions as $K \in \{3, 5, 10, 20, 60\}$.
We also vary the magnitude of the noise as $\sigma \in \{0.05, 0.1, 0.5, 1, 2\}$, which results in a total of 25 instances.

We run the following algorithms:
\begin{itemize}
    \item \bb{TS}: We run Thompson Sampling as our baseline algorithm. All results are stated relative to the performance of TS.
    \item \bb{$\TSUCB$}: We run our algorithm as defined in Section~\ref{sec:linear_bandit} with $m=1$ and $m = 100$, denoted as $\TSUCB(1)$ and $\TSUCB(100)$ respectively.
    \item \bb{Greedy}: We pull the arm that maximizes $\langle \hat{\beta}_k, X_t \rangle$, where $\hat{\beta}_k$ is the posterior mean of $\beta_k$.
    \item \bb{UCB}: We pull the arm with the highest $U_t(a)$ as defined in Section~\ref{sec:linear_bandit} (this is the OFUL algorithm of \cite{abbasi2011improved}).
    \item \bb{IDS}: We run the sample variance-based IDS (Algorithm 6 from \cite{russo2018ids}) with $m = 1000$ samples.
\end{itemize}
All algorithms are given knowledge of the prior for the parameters and the variance of the noise, $\sigma^2$, so that correct posteriors can be computed if the algorithm requires it.

\begin{figure*}[h]
\begin{center}
\subfigure[\TSUCB(1)]{\includegraphics[width=0.3\columnwidth]{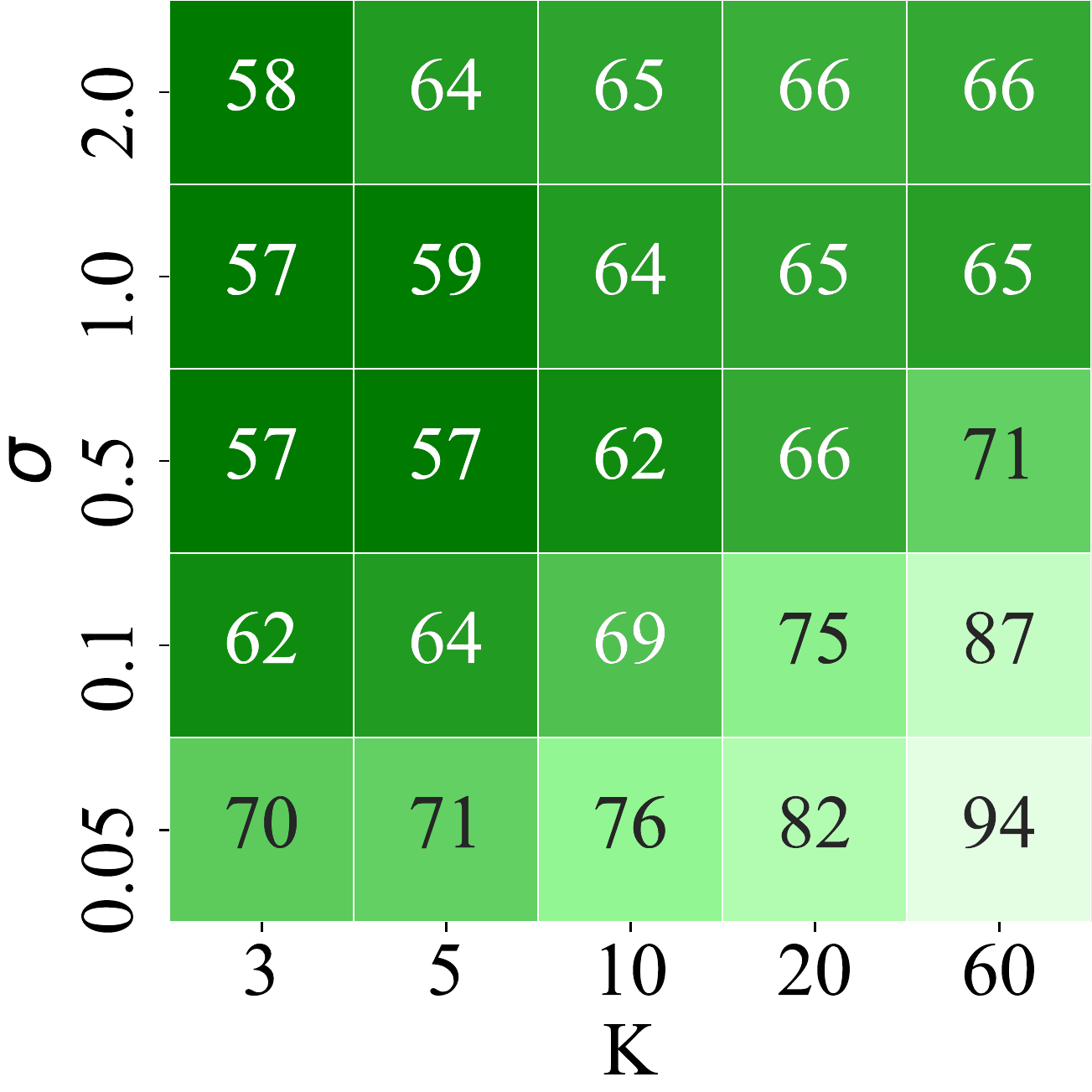}}  \hspace{0.003\textwidth}
\subfigure[\TSUCB(100)]{\includegraphics[width=0.3\columnwidth]{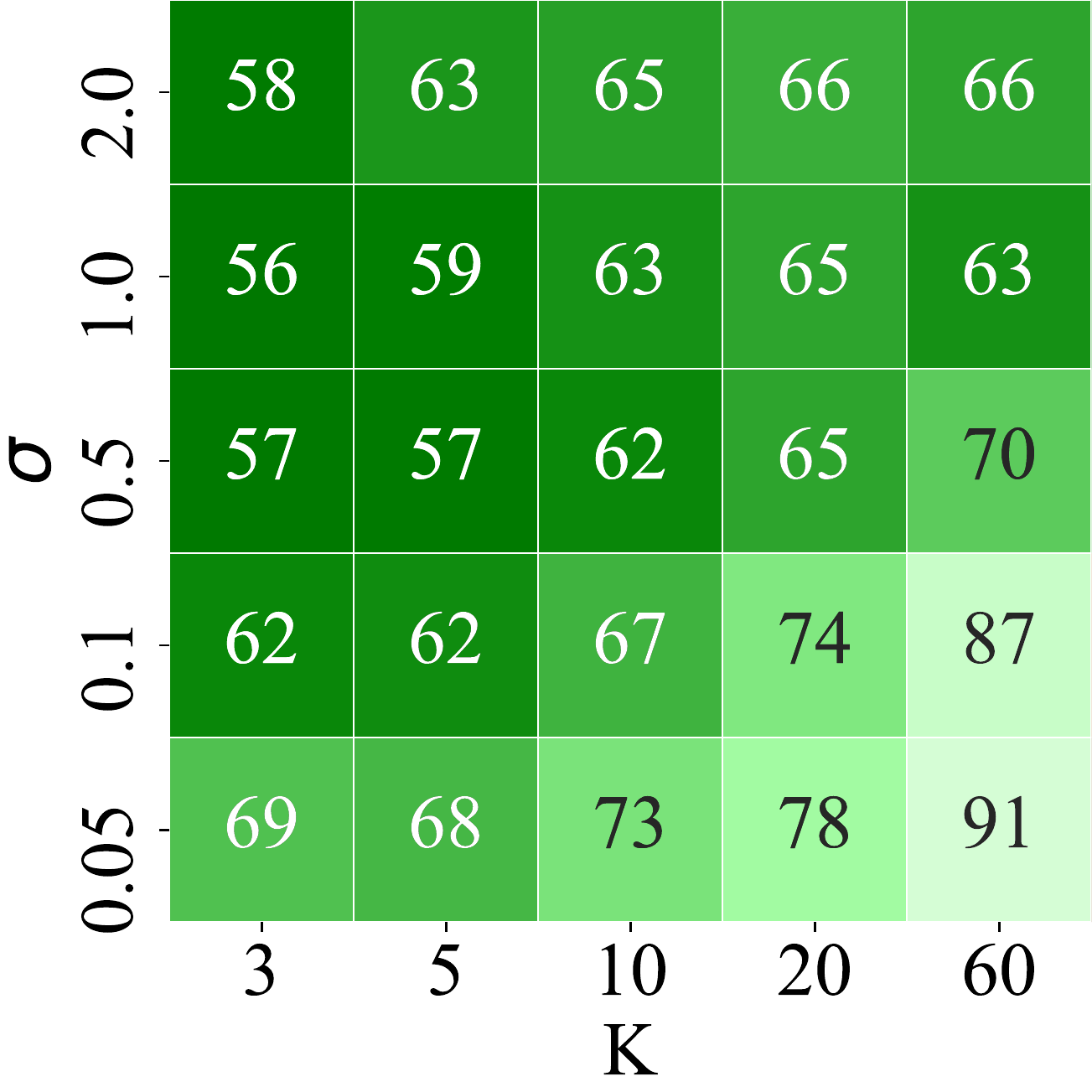}} \\
\subfigure[Greedy]{\includegraphics[width=0.3\columnwidth]{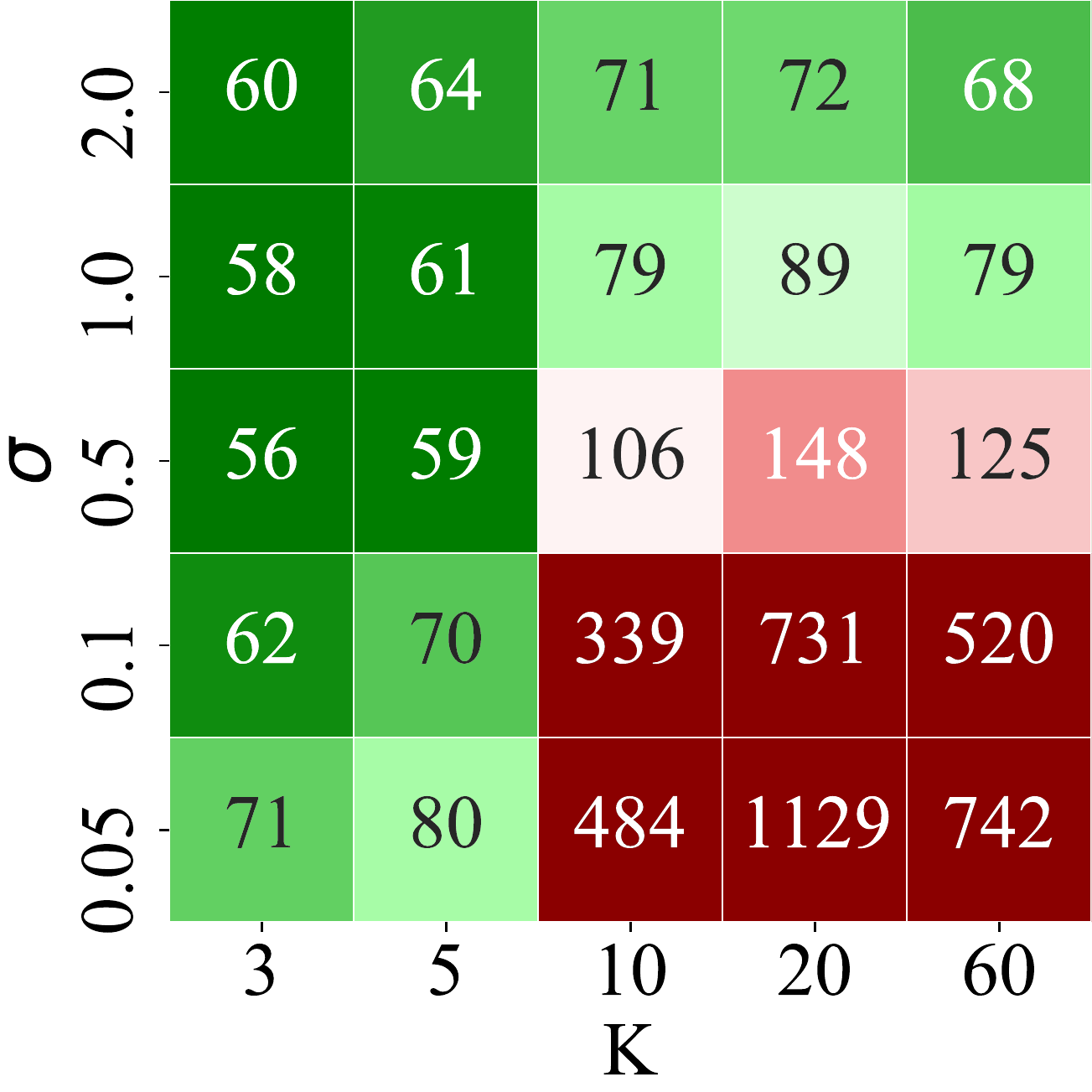}} \hspace{0.003\textwidth}
\subfigure[UCB]{\includegraphics[width=0.3\columnwidth]{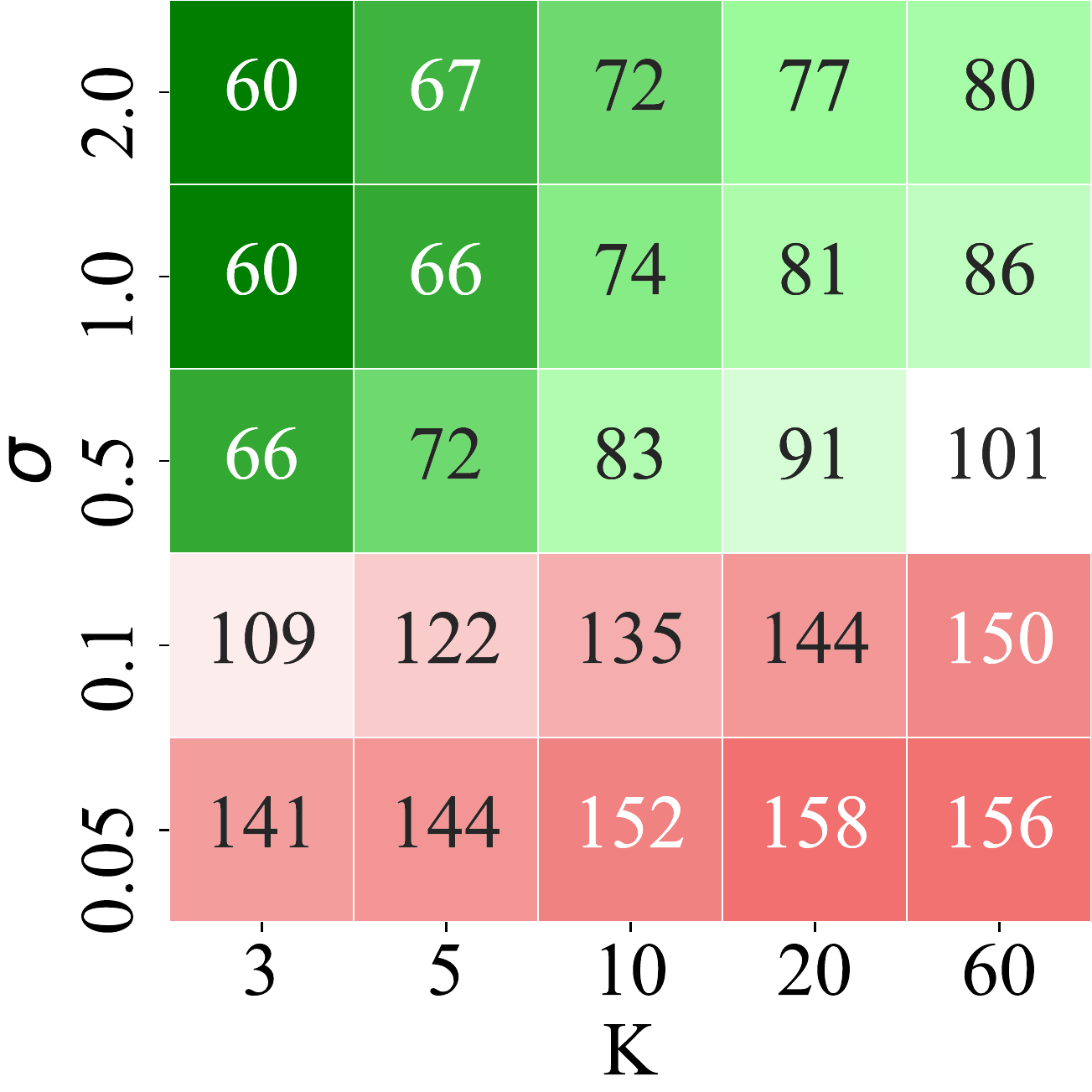}}
\subfigure[IDS]{\includegraphics[width=0.3\columnwidth]{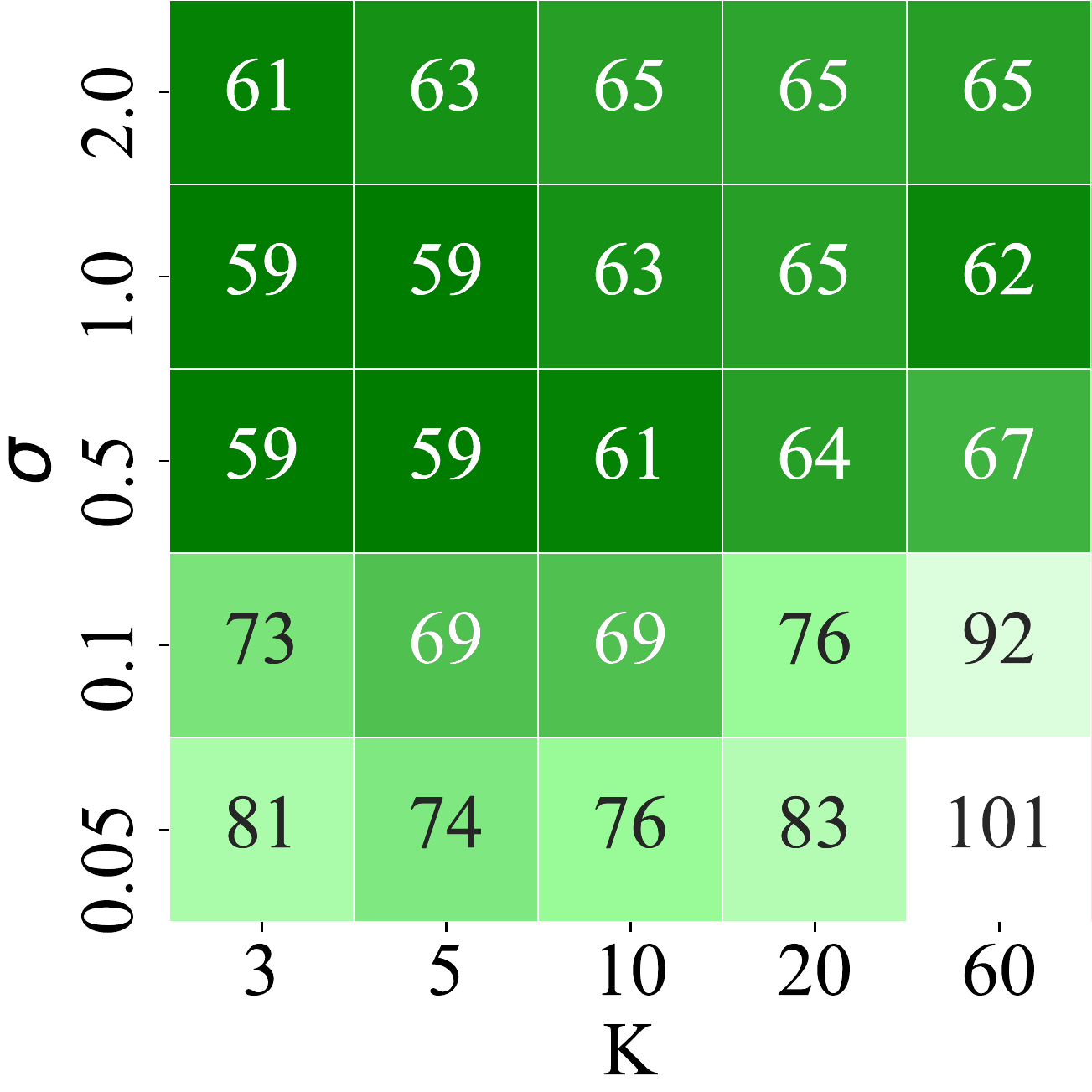}}
\caption{TS-UCB improves on TS across the board.
Grid reports mean regret of each policy as a percentage of regret of Thompson Sampling over 200 runs.
A number smaller than 100 means the regret of that policy is smaller than TS; otherwise the regret is larger than TS.
$\TSUCB(m)$ refers to the algorithm using $m$ samples.}
\label{fig:syntheticlin}
\end{center}
\end{figure*}

For each algorithm and problem instance, we simulate 200 runs over a time horizon of $T = 10,000$.
We report the average regret as a percentage of the regret from the TS policy (that is, we estimate $100 \cdot \bE\left[\frac{\Reg(\text{ALG})}{\Reg(\text{TS})}\right]$).
The results are shown in Figure~\ref{fig:syntheticlin}.

\subsubsection{Synthetic Experiment Results}
We see that both $\TSUCB(1)$ and $\TSUCB(100)$ outperforms TS across the board, almost halving regret in many instances.
The general trend is that $\TSUCB$ has a greater performance improvement over TS when $\sigma$ is higher, which correspond to the ``harder'' instances.
Over the 25 instances, the regret from $\TSUCB(1)$ and $\TSUCB(100)$ was 67.9\% and 66.6\% of the regret of TS respectively.

We see that $\TSUCB(100)$ performs better than $\TSUCB(1)$ overall. $\TSUCB(100)$ outperforms $\TSUCB(1)$ in 21 out of 25 instances, and on average, the regret for $\TSUCB(1)$ is higher by 3.2\% compared to $\TSUCB(100)$.
However, this improvement is small relative to the performance gain over TS; most of the benefit of $\TSUCB$ is captured by using just a single sample.

Both greedy and UCB have inconsistent performances relative to TS.
The greedy algorithm outperforms TS and performs similarly to $\TSUCB$ when the number of actions is small. This is consistent with the result of \cite{bastani2020mostly}, which prove greedy is rate optimal when $K=2$ and the contexts are diverse. However, when $K$ is large, context diversity becomes insufficient to guarantee enough exploration for every arm, resulting in poor performance.
For UCB, we see that performance is poor when the noise is small; this is likely due to UCB being too conservative.
Nonetheless, $\TSUCB$ almost always outperforms both algorithms across the board.



IDS performs well across the board compared to TS.
Its performance is similar to $\TSUCB$ but slightly worse on average --- across the 25 instances, regret for IDS was 5.6\% higher than $\TSUCB(100)$ and 3.8\% higher than $\TSUCB(1)$.
IDS was expected perform well, as it is considered the state-of-the-art algorithm.
However, we see that a much simpler algorithm, $\TSUCB$, performs as well, and often better, than IDS.


\subsection{Personalized News Article Recommendation}




We test the same set of bandit algorithms on a real-world dataset for personalizing news article recommendations for users that land on the front page of the Yahoo! website.
In this setting, when a user goes on the website, the website must choose one article to recommend out of a pool of available articles at that time, in which the user may click on the article to read the full story.
The pool of available articles changes throughout the day.
The goal is to recommend articles that maximize the click-through rate.

We use a dataset that was generated by an experiment done by Yahoo! from 10 days in October 2011, made available through the Yahoo Webscope Program\footnote{\url{https://webscope.sandbox.yahoo.com/}\nocite{yahoowebscope}}.
In the experiment, when a user appeared, the article that was recommended was chosen uniformly at random out of all available at the time, and whether the user clicked on the article was logged.
Each of these users is associated with a context vector of dimension $d=136$ that corresponds to user covariates such as gender and age.
Each sample in the dataset corresponds to the user context, the set of articles available, the article recommended, and whether the user clicked on the article.
There are 1.3 - 2.2 million samples for each of the 10 days.
A very similar dataset was used in \cite{chapelle2011empirical}, which was one of the first papers to display superior empirical performance of Thompson Sampling.

Given this dataset, we use the following method to evaluate a bandit policy.
For each article, we first learn a mapping from user features to their click probabilities using a logistic regression model on the entire dataset.
We only considered articles that had more than 5000 samples so that we could learn an accurate mapping.
We then use this logistic regression model to compute $\hat{p}_{ua}$, an estimate of the probability that a user $u$ will click an article $a$.
We then simulate a bandit policy, where the reward observed from the chosen arm is a Bernoulli random variable with parameter $\hat{p}_{ua}$\footnote{The motivation for this bandit evaluation method was solely to speed up computation for the IDS algorithm. We initially tried the offline evaluation of \cite{li2011unbiased}, which does not require learning a separate regression model; however, running this method once for one day's worth of data using the IDS policy took over 3 days.
Using the logistic regression model reduced the simulation time by more than $20\times$.
}.
Then, the total regret is computed as $\sum_{t=1}^T \left(\max_{a \in \cA_t} \hat{p}_{ua} - \hat{p}_{u A_t}\right)$.
We consider each of the 10 days as separate bandit problem instances, and we also randomly sample 2\% of the dataset so that we have around 25,000-45,000 samples for each problem instance (each ``run'' used a new random sample).
Furthermore, at each time step, out of the articles that were marked to be available in the dataset at that time (this pool contained 20-40 articles), we chose only 10 articles to be available to the bandit algorithm, chosen as the first 10 alphabetically\footnote{Reducing the number of arms was done to also to speed up the computation for IDS, as the runtime of IDS is quadratic in the number of arms.}.

We model this setting as a linear contextual bandit and use the Bayesian structure of \cite{riquelme2018deep}.
Each article corresponds to an arm, and each arm is associated with unknown parameters $\beta_{a} \in \bR^{d}$ and $\sigma_a^2 \in \bR$.
The reward for arm $a$ corresponding to context $X$ is modeled as $Y = \beta_{a}^\top X + \eps$ where $\eps \sim N(0, \sigma_a^2)$.
We model the joint distribution of the parameters $\beta_a$ and $\sigma_a^2$, where we assume they are distributed according to a Gaussian and an Inverse Gamma distribution respectively.
At time $t$, suppose there have been $t_a$ pulls of arm $a$, corresponding to the contexts $\bm{X_t} \in \bR^{t_a \times d}$ and rewards $\bm{Y}_t \in \bR^{t_a}$.

Then, the posterior distributions are $\sigma_a^2 \sim \text{IG}(a_t, b_t)$, and $\beta_a | \sigma_a^2 \sim N(\mu_t, \sigma_a^2 \Sigma_t)$, where
\begin{align}
&\Sigma_t = (\bm{X}_t^{\top}\bm{X}_t + \Lambda_0)^{-1} && \mu_t = \Sigma_t (\Lambda_0 \mu_0 + \bm{X}_t^{\top}\bm{Y}_t), \label{eq:bayesian_structure1} \\
&a_t = a_0 + t_a/2 && b_t = b_0 + \frac{1}{2}(\bm{Y}_t^{\top}\bm{Y}_t + \mu_0 \Sigma_0 \mu_0 - \mu_t^{\top} \Sigma_t^{-1} \mu_t). \label{eq:bayesian_structure2}
\end{align}
We initialize the prior parameters to be $a_0 = b_0 = 6$, $\mu_0 = 0_d$, and $\Lambda_0 = 4 I_d$.
At each time step, for each arm, we first sample $\tilde{\sigma}^2$ from its posterior, and then we use the conditional posterior for $\beta_a$, $N(\mu_t, \tilde{\sigma}^2 \Sigma_t)$, to run the following linear contextual bandit algorithms: TS, $\TSUCB(1)$, $\TSUCB(100)$, IDS, Greedy, and UCB.
For each of the 10 days, we simulated 20 runs for each policy.
We report the regret of each policy as a percentage of the regret of TS, shown in Table~\ref{tab:news}.

\begin{table}[h]
\TableSpaced 
\caption{
Yahoo! article recommendation simulation results from 10 days in October 2011.
TS-UCB provides an improvement over TS across the board.
For each policy, we report the regret as a percentage of regret of Thompson Sampling (with 95\% confidence intervals) for that approach.
For each day, the policy with the lowest average regret is bolded.
IDS requires one thousand samples from the posterior at each time step; TS-UCB(1) and TS-UCB(100) requires one and one hundred samples respectively.
}
\label{tab:news}
\vspace{2mm} 
\begin{center}
\begin{tabular}{lccccc}
\toprule
Day  & $\TSUCB(1)$& $\TSUCB(100)$ & $\IDS$ & Greedy & UCB \\
\midrule
1 & $91.0 \pm 1.5$ & $\bm{89.1 \pm 1.6}$ & $90.0 \pm 1.5$ & $106.1 \pm 6.2$ & $106.2 \pm 1.1$ \\
2 & $86.3 \pm 1.3$ & $\bm{82.2 \pm 1.9}$ & $84.8 \pm 2.2$ & $100.6 \pm 9.8$ & $121.9 \pm 1.7$ \\
3 & $85.8 \pm 1.9$ & $\bm{84.1 \pm 1.4}$ & $84.8 \pm 1.7$ & $122.3 \pm 7.0$ & $124.9 \pm 1.6$ \\
4 & $92.5 \pm 1.7$ & $91.6 \pm 2.0$ & $\bm{90.9 \pm 1.6}$ & $107.0 \pm 6.2$ & $123.8 \pm 2.1$ \\
5 & $91.1 \pm 1.8$ & $\bm{89.8 \pm 1.7}$ & $90.9 \pm 1.7$ & $100.7 \pm 3.2$ & $110.7 \pm 1.4$ \\
6 & $85.1 \pm 1.4$ & $83.7 \pm 0.7$ & $\bm{83.2 \pm 1.2}$ & $105.6 \pm 4.4$ & $102.2 \pm 1.1$ \\
7 & $96.2 \pm 1.5$ & $96.3 \pm 1.9$ & $94.0 \pm 2.2$ & $\bm{88.5 \pm 7.3}$ & $121.8 \pm 1.2$ \\
8 & $90.7 \pm 2.4$ & $\bm{89.5 \pm 2.1}$ & $90.0 \pm 2.4$ & $106.9 \pm 4.3$ & $119.8 \pm 2.0$ \\
9 & $92.3 \pm 1.7$ & $\bm{88.8 \pm 2.4}$ & $90.4 \pm 2.0$ & $92.4 \pm 7.7$ & $116.4 \pm 2.1$ \\
10 & $88.1 \pm 1.9$ & $\bm{86.4 \pm 3.0}$ & $86.7 \pm 1.3$ & $93.1 \pm 5.9$ & $122.8 \pm 1.6$ \\
\midrule
Overall & $89.9 \pm 0.7$ & $\bm{88.2 \pm 0.8}$ & $88.6 \pm 0.7$ & $102.3 \pm 2.4$ & $117.1 \pm 1.2$ \\
\bottomrule
\end{tabular}
\end{center}
\end{table}

\subsubsection{Personalized News Article Results}
$\TSUCB(100)$ performed the best overall, having the lowest average regret in 7 out of 10 days.
We see that $\TSUCB(1)$, $\TSUCB(100)$ and IDS significantly outperform TS in all instances --- reducing regret by more than 10\% on average.
The relative performance of these three algorithms are comparable, where $\TSUCB(100)$ slightly outperforms IDS, and IDS slightly outperforms $\TSUCB(1)$.
Greedy performs similarly to TS, while UCB is clearly outperformed by TS.
Overall, we see a similar pattern in performance as compared to the synthetic experiments;
$\TSUCB$ clearly outperforms TS, and moreover, often outperforms IDS while being much cheaper than IDS computationally.

\subsection{Deep Bandit Benchmark} \label{sec:exp_real_world}

In challenging bandit models such as the deep contextual bandit discussed in Example~\ref{ex:neural_linear}, computing a posterior is challenging.
\cite{riquelme2018deep} evaluate a large number of posterior approximation methods on a variety of real-world datasets for such a  contextual bandit problem.
Their results suggest that performing posterior sampling using the ``Neural Linear'' method, described in Example~\ref{ex:neural_linear},
 is an effective and robust approach.
We evaluate $\TSUCB$ on the benchmark problems in \cite{riquelme2018deep} and compare its performance to TS, IDS, greedy, and UCB.


For a finite action set of size $K$, Neural Linear maintains one neural network, $h_t:\bR^{d'} \rightarrow \bR^{d}$, as well as posterior distributions on $K$ parameter vectors $\beta_a \in \bR^d$ and $K$ scalar parameters $\sigma_a^2 \in \bR$.
At time $t$, the posteriors on $\beta_a$ and $\sigma_a^2$ are computed {\em ignoring the uncertainty} in the estimate of $h_t(\cdot)$
so that this computation is equivalent to bayesian linear regression.
Specifically, we assume a linear contextual bandit model on the context \ii{representation} $h_t(X)$. We use the same Bayesian structure as in \cite{riquelme2018deep}, described by \eqref{eq:bayesian_structure1}-\eqref{eq:bayesian_structure2}; the only difference is that the matrix $\bm{X}_t$ is replaced by a matrix whose $i$'th row is $h_t(X_i)$ instead of $X_i$.

While the original dimension, $d'$, varies across datasets, the last layer of the neural network has same dimension $d=50$ for every dataset.
We use a neural network with two fully connected layers of size 50 for $h_t(\cdot)$.
The network is updated every 50 time steps, in which the network minimizes mean squared error for the observed rewards using the RMSProp optimizer \citep{hinton2012neural}.

We replicate the experiments from \cite{riquelme2018deep} with the same real-world datasets.
These datasets vary widely in their properties;
see Appendix A of \cite{riquelme2018deep} for the details of each dataset.
We simulate 200 runs for each dataset and algorithm.
For each dataset, one ``run'' is defined as 10,000 data points randomly drawn from the entire dataset;
that is, there are 10,000 time steps\footnote{The financial dataset did not have 10,000 data points, so we used $3,000$ data points for this dataset only.}, and each data point (or ``context'') arrives sequentially in a random order.
We report the regret of each policy as a percentage of the regret of TS, shown in Table~\ref{tab:all}.

\begin{table}[h]
\TableSpaced 
\caption{Deep Bandit benchmark \cite{riquelme2018deep} results for the Neural Linear posterior approximation method.
For each posterior approximation approach, the regret is reported as a percentage of regret of Thompson Sampling (with 95\% confidence intervals) for that approach.
For each dataset, the policy with the lowest average regret is bolded.
}
\label{tab:all}
\vspace{2mm} 
\centering
\begin{tabular}{lccccccc}
\toprule
Dataset   & $d'$ & $K$& $\TSUCB(1)$& $\TSUCB(100)$ & $\IDS$ & Greedy & UCB \\
\midrule
adult & 14 & 86  & $98.8 \pm 0.2$ & $\bm{98.6 \pm 0.2}$ & $\bm{98.6 \pm 0.2}$ & $103.4 \pm 0.7$ & $101.0 \pm 0.2$ \\
census & 369 & 9  & $99.4 \pm 0.5$ & $99.2 \pm 0.5$ & $99.2 \pm 0.5$ & $\bm{92.2 \pm 0.5}$ & $105.2 \pm 0.5$ \\
covertype & 54 & 7  & $98.7 \pm 0.6$ & $98.6 \pm 0.6$ & $98.4 \pm 0.6$ & $\bm{91.5 \pm 0.6}$ & $110.3 \pm 0.5$ \\
financial & 21 & 8  & $60.0 \pm 0.9$ & $\bm{54.7 \pm 0.7}$ & $56.3 \pm 0.8$ & $101.6 \pm 9.8$ & $160.7 \pm 2.3$ \\
jester & 32 & 8  & $99.4 \pm 0.3$ & $99.2 \pm 0.3$ & $99.6 \pm 0.3$ & $\bm{96.6 \pm 0.4}$ & $113.1 \pm 0.8$ \\
mushroom & 117 & 2  & $108.0 \pm 11.1$ & $\bm{98.6 \pm 5.1}$ & $118.6 \pm 15.7$ & $189.5 \pm 32.4$ & $918.3 \pm 53.0$ \\
statlog & 9 & 7  & $89.7 \pm 0.6$ & $74.9 \pm 0.6$ & $\bm{73.9 \pm 0.6}$ & $317.5 \pm 27.8$ & $322.9 \pm 2.0$ \\
\bottomrule
\end{tabular}
\end{table}


\subsubsection{Deep Bandit Benchmark Results}

\cite{riquelme2018deep} establish TS along with the neural linear approach to posterior sampling as a benchmark algorithm for deep contextual bandits.
We see here that $\TSUCB$ improves upon TS on every dataset except possibly mushroom, and it offers significant improvements in datasets financial and statlog.
Similarly to the synthetic experiments, $\TSUCB(100)$ always outperforms $\TSUCB(1)$.

The performance of the other algorithms relative to both TS and $\TSUCB$ is also similar to the results of the synthetic experiments. IDS usually outperforms TS, and has a similar performance to $\TSUCB$ but slightly worse in some cases.
On average, the regret for IDS was 4.1\% higher than $\TSUCB(100)$ and 0.02\% higher than $\TSUCB(1)$.
Greedy has a very inconsistent performance across datasets.
It outperforms all other algorithms in three datasets (census, covertype, jester), suggesting that no exploration is needed in these cases.
However, its poor performance in mushroom and statlog suggest that exploration is indeed necessary in several real-world settings.
UCB is consistently outperformed by both TS and $\TSUCB$.






In summary, both the synthetic and real-world experiments suggest the same conclusion:
$\TSUCB$ outperforms TS across a comprehensive suite of experiments with essentially no additional computation.
Moreover, $\TSUCB$ consistently matches or improves upon the state-of-the-art algorithm IDS, while being a much simpler algorithm than IDS both computationally and conceptually.

\section{Outline of Regret Analysis} \label{sec:pfsketch}

In this section, we give an outline of the proofs of Theorem~\ref{thm:mainK} and \ref{thm:mainlinear}. The full proofs can be found in Appendix~\ref{app:A}.

We first state two known results results on upper bounding $\sumt \bE[\rad_t(A_t)]$ for the two bandit settings that follow from standard UCB analyses.
For the $K$-armed setting, the proof of Proposition 2 of \cite{russo2014learning} implies the following result.
\begin{theorem} \label{thm:sqrtbound}
For the $K$-armed bandit, using the UCBs as defined in \eqref{eq:karmucb},
\begin{align*}
\sumtK \bE[\rad_t(A_t)] \leq 2 \sqrt{3KT \log T },
\end{align*}
for any sequence of actions $A_t$.
\end{theorem}

Similarly, in the linear bandit setting, the proof of Theorem 3 of \cite{abbasi2011improved} (using the parameters $\delta=T^{-3}, \lambda=1$) implies the following result.
\begin{theorem} \label{thm:sqrtbound_linear}
For the linear bandit, using the UCBs as defined in \eqref{eq:linearucb},
\begin{align*}
\sumt \bE[\rad_t(A_t)] \leq& 4\sqrt{Td \log(1+TL/d)}(S+ \subg \sqrt{6 \log(T) + d \log(1 + T/d)}) = O(d \log T \sqrt{T})
\end{align*}
for any sequence of actions $A_t$.
\end{theorem}

Next, it is useful to extend the definition of $\Psi_t$ to randomized actions.
If $\nu$ is a probability distribution over $\cA$, define
\begin{align}
\bPsi_t(\nu) \triangleq \frac{  \tf_t - \bE_{A_t \sim \nu}[\hmu_{t}(A_t)] }{\bE_{A_t \sim \nu}[\rad_t(A_t)]}. \label{eq:defratio2}
\end{align}
Using this definition, we show (Lemma~\ref{lem:random}) that for any policy $(\pi_t)_{t \geq 1}$, surely,
\begin{align} \label{eq:tsucb_less_randomized}
\Psie_t(A^{\TSUCB}_t) \leq \bPsi_t(\pi_t).
\end{align}

Now, assume the following two approximations hold at every time step:
\begin{enumerate}[label=(\roman*)]
  \item \label{enum1} $\tf_t$ approximates the expected optimal reward: $\tf_t \approx \bE[\ft(\ssA) | H_t]$.
  \item \label{enum2} $\hmu_t(a)$ approximates the expected reward of action $a$: $\hmu_t(a) \approx \bE[\ft(a)|H_t]$.
\end{enumerate}

The Bayes regret for $\TSUCB$ can be decomposed as
\begin{align}
\BReg(T, \pi^{\TSUCB}) &= \sumt \bE[\bE[\ft(\ssA_t) - \ft(A^{\TSUCB}_t)|H_t]] \nonumber \\
            &\approx \sumt \bE[\tf_t  - \hmu_t(A^{\TSUCB}_t) ] \nonumber \\
            &= \sumt \bE\left[\Psie_t(A^{\TSUCB}_t) \rad_t(A^{\TSUCB}_t) \right], \label{eq:sketch}
\end{align}
where the second step uses \ref{enum1}-\ref{enum2}, and the third step uses the definition \eqref{eq:defratio1}.

\eqref{eq:sketch} decomposes the regret into the product of two terms: the ratio $\Psie_t(A^{\TSUCB}_t)$ and the radius of the action taken.
For the second piece, standard analyses for the UCB algorithm found in the literature bound regret by bounding the sum $\sumt \bE[\rad_t(A_t)]$ for any sequence of actions $A_t$.
Therefore, if $\Psie_t(A^{\TSUCB}_t)$ can be upper bounded by a constant, the regret bounds found for UCB can be directly applied.

We show $\bPsi_t(\pi_t^{\TSalg}) \lessapprox 1$, where $\TSalg$ is the Thompson Sampling policy (this is stated formally and shown in Lemma~\ref{prop:boundratio}.).
In light of \eqref{eq:tsucb_less_randomized}, this implies $\Psie_t(A^{\TSUCB}_t) \lessapprox 1$.
Plugging this back into \eqref{eq:sketch} gives us
$\BReg(T, \pi^{\TSUCBE}) \lessapprox  \sumt \bE\left[\rad_t(A^{\TSUCB}_t)\right]$, which lets us apply UCB regret bounds from the literature and finishes the proof.

This method of decomposing the regret into the product of two terms (as in \eqref{eq:sketch}) and minimizing one of them was used in \cite{russo2018ids} for the IDS policy.
The optimization problem in IDS is difficult, as the term that is minimized involves evaluating the information gain, requiring computing integrals over high-dimensional spaces.
The optimization problem for $\TSUCB$ is almost trivial, but it trades off on the ability to incorporate complicated information structures as IDS can.

The above proof outline can be used to prove the following proposition.
\begin{proposition} \label{thm:maingenerale}
Suppose $\rad_t(a) \in [\rmin, \rmax]$ for all $a \in \cA$ and $t \geq 1$.
Using the UCBs as defined in \eqref{eq:karmucb} for the $K$-armed bandit, and \eqref{eq:linearucb} for the linear bandit,
\begin{align} \label{eq:prop1}
 \BReg(T, \pi^{\TSUCB}) &\leq 2 \sumt \bE[\rad_t(\At)] + \frac{\rmax}{\rmin} \left(1+ \frac{2T}{\sqrt{m}} \right) + T^{-2}.
\end{align}
\end{proposition}

The approximation $\tf_t \approx \bE[\ft(\ssA) | H_t]$ used in the proof sketch only holds when $m$ is large; the fact that this doesn't hold contributes to the $\frac{1}{\sqrt{m}}$ term in \eqref{eq:prop1}, which goes to zero as $m \rightarrow \infty$.
To cover the case when $m$ is small, we also show the following proposition, which has the opposite relationship with respect to $m$.

\begin{proposition} \label{thm:maingeneral}
Using the UCBs as defined in \eqref{eq:karmucb} for the $K$-armed bandit, and \eqref{eq:linearucb} for the linear bandit,
\begin{align*}
\BReg(T, \pi^{\TSUCB}) \leq  2 \sumt \bE[\rad_t(\At)] + (m+1)T^{-2}.
\end{align*}
\end{proposition}

The final step of showing Theorems~\ref{thm:mainK} and \ref{thm:mainlinear} involves combining these two propositions to remove the dependence on $m$ and plugging in the known bounds for $\sumt \bE[\rad_t(A_t)] $ from Theorems~\ref{thm:sqrtbound} and \ref{thm:sqrtbound_linear}.
The formal proofs of the theorems can be found in Appendix~\ref{app:A}.

%% file: sections/regret_analysis.tex
For our analysis, we introduce lower confidence bounds $(L_t)_{t \geq 1}$, which we define in a symmetric way to upper confidence bounds:
$L_t(a) \triangleq \hmu_t(a) - (U_t(a) - \hmu_t(a))$.

We first state a lemma that says that the confidence bounds are valid with high probability
\begin{lemma} \label{lem:hoeffding_general}
Using the functions $\{\hmu_t\}_{t \geq 1}$, $\{U_t\}_{t \geq 1}$ as defined in \eqref{eq:karmucb} in the $K$-armed setting and \eqref{eq:linearucb} in the linear bandit setting,
for any $t \leq T$, $\Pr(\ft(A) < U_t(A)) \leq T^{-3}$, where $A$ is any deterministic or random action.
The analogous bounds hold for lower confidence bounds, i.e. $\Pr(\ft(A) > L_t(A)) \leq T^{-3}$.
\end{lemma}
For completeness, the proof of Lemma~\ref{lem:hoeffding_general} can be found in \ref{subsub:provelemmas}.
The following corollary is immediate using the law of total expectation and the fact that $\ft(A) \geq -1$.
\begin{corollary} \label{corr:lcb}
For any $t \leq T$,
$\bE[- \ft(A)] \leq \bE[- L_t(A)] + T^{-3}$, where $A$ is any deterministic or random action.
\end{corollary}

The next two subsections prove Proposition~\ref{thm:maingenerale} and \ref{thm:maingeneral} respectively.
The final step of the proof combines these propositions with the known bounds for $\sumt \bE[\rad_t(A_t)]$ from Theorems~\ref{thm:sqrtbound} and \ref{thm:sqrtbound_linear}, and can be found in \ref{sec:combine}.

\subsection{Proof of Proposition~\ref{thm:maingenerale}.}
We first state the result claimed in \eqref{eq:tsucb_less_randomized}
whose proof is deferred to \ref{subsub:provelemmas}.
\begin{lemma} \label{lem:random}
For any distribution $\tau$ over $\cA_t$,
 $\Psi_t(\At) \leq \bPsi_t(\tau)$ almost surely.
\end{lemma}
Next, we upper bound the ratio $\Psi_t(\At)$ by analyzing the Thompson Sampling policy.
\begin{lemma} \label{prop:boundratio}
$\Psi_t(\At) \leq 1 + \frac{1}{\rmin}(\Pr(f_{\theta}(\ssA) > U_t(\ssA) | H_t) +  \tf_t - \bE[f_{\theta}(\ssA)|H_t])$ almost surely.
Equivalently, using \eqref{eq:defratio1},
\begin{align} \label{eq:boundratio}
\tf_t - \hmu_{t}(\At)
\leq \rad_t(\At)(1 + \frac{1}{\rmin}(\Pr(f_{\theta}(\ssA) > U_t(\ssA) | H_t) +  \tf_t - \bE[f_{\theta}(\ssA)|H_t])).
\end{align}
\end{lemma}
\begin{myproof}[Proof.]
Let $\pi^{\TSalg}$ be the Thompson sampling policy.
We show the inequality for $\bPsi_t(\pi_t^{\TSalg})$ instead, and then use $\Psi_t(\At) \leq \bPsi_t(\pi_t^{\TSalg})$ from Lemma~\ref{lem:random} to get the desired result.

By definition of TS,
$\pi_t^{\TSalg} = \pi_t^{\TSalg}(H_t)$ is the distribution over $\cA_t$ corresponding to the posterior distribution of $\ssA$ conditioned on $H_t$.
Then, if $A_t$ is the action chosen by TS at time $t$,
we have $\bE[U_t(A_t) |H_t] = \bE[U_t(\ssA)|H_t]$ and $\bE[\hmu_t(A_t) |H_t] = \bE[\hmu_t(\ssA)|H_t]$.
Using this, we can write $\bPsi_t(\pi_t^{\TSalg})$ as
\begin{align} \label{eq:psi_ts}
\bPsi_t(\pi_t^{\TSalg}) = \frac{\tf_t - \bE[\hmu_{t}(A_t)|H_t]}{\bE[U_t(A_t) - \hmu_{t}(A_t)|H_t]}
= \frac{\tf_t - \bE[\hmu_{t}(\ssA)|H_t]}{\bE[U_t(\ssA) - \hmu_{t}(\ssA)|H_t]}.
\end{align}

By conditioning on the event $\{f_{\theta}(\ssA) \leq U_t(\ssA)\}$, the following inequality follows from the fact that $f_{\theta}(\ssA) \leq 1$.
\begin{align} \label{eq:boundratiopf}
\bE[f_{\theta}(\ssA) | H_t] \leq \bE[U_t(\ssA)|H_t] + \Pr(f_{\theta}(\ssA) > U_t(\ssA) | H_t).
\end{align}
Consider the numerator of \eqref{eq:psi_ts}. We add and subtract $\bE[f_{\theta}(\ssA)|H_t]$ and use \eqref{eq:boundratiopf}:
\begin{align}
\tf_t - \bE[\hmu_{t}(\ssA)|H_t] &= \bE[f_{\theta}(\ssA)-\hmu_{t}(\ssA)|H_t]  + \tf_t - \bE[f_{\theta}(\ssA)|H_t] \nonumber \\
 &\leq \bE[U_t(\ssA) - \hmu_{t}(\ssA)|H_t] + \Pr(f_{\theta}(\ssA) > U_t(\ssA) | H_t) +  \tf_t - \bE[f_{\theta}(\ssA)|H_t]. \label{eq:boundratiopf2}
\end{align}
The first term of \eqref{eq:boundratiopf2} is equal to the denominator of $\bPsi_t(\pi^{\TSalg})$.
Therefore,
\begin{align*}
 \bPsi_t(\pi^{\TSalg})
             &\leq 1 + \frac{\Pr(f_{\theta}(\ssA) > U_t(\ssA) | H_t) +  \tf_t - \bE[f_{\theta}(\ssA)|H_t]}{\bE[U_t(\ssA) - \hmu_{t}(\ssA)|H_t]} \\
             &\leq 1 + \frac{1}{\rmin}(\Pr(f_{\theta}(\ssA) > U_t(\ssA) | H_t) +  \tf_t - \bE[f_{\theta}(\ssA)|H_t]).
\end{align*}
\end{myproof}


The next lemma simplifies the expectation of \eqref{eq:boundratio} using Cauchy-Schwarz.

\begin{lemma}\label{lem:mainlem}
For any $t$, $\bE[\tf_t - \hmu_{t}(\At)] \leq \bE[\rad_t(\At)]  + \frac{\rmax}{\rmin} \left(\frac{1}{T} + \frac{2}{\sqrt{m}} \right)$.
\end{lemma}

\begin{myproof}[Proof.]

Taking the expectation of \eqref{eq:boundratio} gives us
\begin{align}
&\bE[\tf_t - \hmu_{t}(\At)] \nonumber \\
\leq& \bE[\rad_t(\At)(1 + \frac{1}{\rmin}(\Pr(f_{\theta}(\ssA) > U_t(\ssA) | H_t) +  \tf_t - \bE[f_{\theta}(\ssA)|H_t]))] \nonumber \\
=& \bE[\rad_t(\At)] \nonumber \\
 &+\frac{1}{\rmin}\bE[\rad_t(\At)\Pr(f_{\theta}(\ssA) > U_t(\ssA) | H_t)] \label{eq:firstcs} \\
 & +  \frac{1}{\rmin}\bE[\rad_t(\At) (\tf_t - \bE[f_{\theta}(\ssA)|H_t])]. \label{eq:secondcs}
\end{align}

We will now upper bound \eqref{eq:firstcs} and \eqref{eq:secondcs} with $\frac{\rmax}{\rmin} \cdot \frac{1}{T}$ and $\frac{\rmax}{\rmin} \cdot \frac{2}{\sqrt{m}}$ respectively, in which case the result will follow.
First, consider \eqref{eq:firstcs}. Using Cauchy-Schwarz yields
\begin{align}
 &\frac{1}{\rmin} \bE[\rad_t(\At)\Pr(f_{\theta}(\ssA) > U_t(\ssA) | H_t)]  \nonumber \\
 \leq& \frac{1}{\rmin}  \sqrt{\bE[\rad_t(\At)^2] \bE[\Pr(\ft(\ssA) > U_t(\ssA) | H_t)^2]} \nonumber \\
 \leq& \frac{1}{\rmin T}\sqrt{\bE[\rad_t(\At)^2]} \nonumber \\
 \leq& \frac{1}{T} \cdot \frac{\rmax}{\rmin}, \label{eq:cs1}
\end{align}
where the second step uses the following.
\begin{align}
\bE[\Pr(\ft(\ssA) > U_t(\ssA) | H_t)^2] &= \bE[\bE[\bI(\ft(\ssA) > U_t(\ssA)) | H_t]^2] \nonumber \\
&\leq \bE[\bE[\bI(\ft(\ssA) > U_t(\ssA))^2 | H_t]] \nonumber \\
 &=\bE[\bE[\bI(\ft(\ssA) > U_t(\ssA)) | H_t]] \nonumber \\
&\leq  \Pr(\ft(\ssA) > U_t(\ssA)) \nonumber \\
&\leq \frac{1}{T^2}, \nonumber
\end{align}
where the first inequality uses Jensen's inequality, and the last inequality uses Lemma~\ref{lem:hoeffding_general}.

Similarly, we apply Cauchy-Schwarz to \eqref{eq:secondcs}.
\begin{align}
 \frac{1}{\rmin}\bE[\rad_t(\At) (\tf_t - \bE[f_{\theta}(\ssA)|H_t])] &\leq \frac{1}{\rmin}\sqrt{\bE[\rad_t(\At)^2] \bE[(\tf_t - \bE[f_{\theta}(\ssA)|H_t])^2]}. \label{eq:cs2mid}
\end{align}

Recall that $\tf_t = \frac{1}{m} \sum_{i = 1}^m \tf_i$, and $\tf_i$ has the same distribution as $f_{\theta}(\ssA)$ conditioned on $H_t$.
Therefore, $\bE[\tf_t|H_t] = \bE[f_{\theta}(\ssA)|H_t]$. Then, we have
\begin{align}
\bE[(\tf_t - \bE[f_{\theta}(\ssA)|H_t])^2] &= \bE[\bE[(\tf_t - \bE[f_{\theta}(\ssA)|H_t])^2|H_t]] \nonumber \\
&= \bE[\var(\tf_t | H_t)] \nonumber \\\
&= \bE[\frac{1}{m}\var(\tf_i | H_t)] \nonumber  \\\
&\leq \frac{4}{m}. \nonumber 
\end{align}
The last inequality follows since $\tf_i \in [-1, 1]$.
Combining this with \eqref{eq:cs2mid}, we get
\begin{align}
 \frac{1}{\rmin}\bE[\rad_t(\At) (\tf_t - \bE[f_{\theta}(\ssA)|H_t])] &\leq \frac{2}{\rmin\sqrt{m}} \sqrt{\bE[\rad_t(\At)^2]} \nonumber \\
                            &\leq \frac{2}{\sqrt{m}} \cdot \frac{\rmax}{\rmin} \label{eq:cs2}
\end{align}

Substituting \eqref{eq:cs1} and \eqref{eq:cs2} into \eqref{eq:secondcs} yields the desired result.
\end{myproof}

\begin{myproof}[Proof of Proposition~\ref{thm:maingenerale}.]
Conditioned on $H_t$, the expectation of $\ft(\ssA)$ and $\tf_t$ is the same, implying $\bE[\ft(\ssA)] = \bE[\tf_t]$ for any $t$. Therefore, the Bayes regret can be written as $\sumt \bE[\tf_t- \ft(\At)]$.
By adding and subtract $\hmu_t(\At)$, we derive
\begin{align}
 \BReg(T, \pi^{\TSUCBE}) =& \sumt \bE[\tf_t - \hmu_{t}(\At)] + \sumt \bE[\hmu_{t}(\At)- \ft(\At)]. \label{eq:decomp1}
\end{align}


The first sum in \eqref{eq:decomp1} can be bounded by $\sumt \bE[\rad_t(\At)]  + \frac{\rmax}{\rmin} \left(1 + \frac{2T}{\sqrt{m}} \right)$ using Lemma~\ref{lem:mainlem}.
Using Corollary~\ref{corr:lcb}, the second sum in \eqref{eq:decomp1} can be bounded by
$\sumt (\bE[\hmu_{t}(\At)- L_t(\At)] +  T^{-3}) \leq \sumt \bE[\rad_t(\At)] + T^{-2}$.
Substituting these two bounds results in
\begin{align*}
 \BReg(T, \pi^{\TSUCBE}) &\leq 2 \sumt \bE[\rad_t(\At)] + \frac{\rmax}{\rmin} \left(1+ \frac{2T}{\sqrt{m}} \right) + T^{-2}
\end{align*}
as desired.
\end{myproof}

\subsection{Proof of Proposition~\ref{thm:maingeneral}.}

The main idea of this proof is captured in the following lemma, which says that we can essentially replace the term $\bE[f_\theta(\ssA)]$ with $\bE[U_{t}(\At)]$.
\begin{lemma} \label{lem:ucbratioone2}
For every $t$, $\bE[f_\theta(\ssA)] \leq \bE[U_{t}(\At)] + mT^{-3}$.
\end{lemma}
\begin{myproof}
Fix $t$, $H_t$, and $\tf_t$.
For an action $a \in \cA_t$,
if $U_{t}(a) \geq \tf_t$, then $\Psi_{t}(a) \leq 1$ since the denominator of the ratio is always positive.
Otherwise, if $U_t(a) < \tf_t$, then $\Psi_t(a) > 1$.
This implies that an action whose UCB is higher than $\tf_t$ will always be chosen over an action whose UCB is smaller than $\tf_t$.
Therefore, in the case that $\tf_t \leq \max_{a \in \cA_t} U_t(a)$, it will be that $U_t(\At) \geq \tf_t$.
Since $\tf_t \leq 1$, we have
\begin{align*}
\bE[\tf_t |  H_t ] \leq \;&U_t(\At) \Pr(\tf_t\leq \max_{a \in \cA_t} U_t(a) | H_t ) + \Pr(\tf_t> \max_{a \in \cA_t} U_t(a) | H_t ) \\
              \leq \; & U_t(\At) + \Pr(\tf_t > \max_{a \in \cA_t} U_t(a) | H_t ) .
\end{align*}
Since $\tf_t = \frac{1}{m} \sum_{i = 1}^m \tf_i$, if $\tf_t$ is larger than $\max_{a \in \cA_t} U_t(a)$, it must be that at least one of the elements $\tf_i$ is larger than $\max_{a \in \cA_t} U_t(a)$.
Then, the union bound gives us $\Pr(\tf_t > \max_{a \in \cA_t} U_t(a) | H_t ) \leq \sum_{i = 1}^m \Pr(\tf_i > \max_{a \in \cA_t} U_t(a) | H_t )$.
By definition of $\tf_i$, the distribution of $\tf_i$ and $\ft(\ssA_t)$ are the same conditioned on $H_t$.
Therefore,
\begin{align*}
\bE[\tf_t |  H_t ] \leq  U_t(\At) + m \Pr(\ft(\ssA_t) > \max_{a \in \cA_t} U_t(a) | H_t ).
\end{align*}

Using the fact that $\bE[\tf_t | H_t] = \bE[\ft(\ssA_t)|H_t]$ and taking expectations on both sides, we have
\begin{align*}
\bE[\ft(\ssA_t)] &\leq \bE[U_t(\At)] + m \Pr(\ft(\ssA_t) > \max_{a \in \cA_t} U_t(a)) \\
         &\leq \bE[U_t(\At)] + m \Pr(\ft(\ssA_t) >  U_t(\ssA_t)) \\
         &\leq \bE[U_t(\At)] + mT^{-3}.
\end{align*}
The last inequality uses Lemma~\ref{lem:hoeffding_general}.
\end{myproof}


\begin{myproof}[Proof of Proposition~\ref{thm:maingeneral}.]
\begin{align*}
 \BReg(T, \pi^{\TSUCB}) &= \sumt \bE[\ft(\ssA_t) - \ft(\At)] \\
              &\leq \sumt (\bE[U_t(\At) - \ft(\At)]  + mT^{-3}) \\
              &\leq \sumt (\bE[U_t(\At) - L_t(\At)] + T^{-3})  + mT^{-2}\\
              &= 2 \sumt \bE[\rad_t(\At)]  + (m+1)T^{-2},
\end{align*}
where the first inequality uses Lemma~\ref{lem:ucbratioone2} and the second inequality uses Corollary~\ref{corr:lcb}.
\end{myproof}

\subsection{Final step of proof.} \label{sec:combine}

\begin{myproof}[Proof of Theorem~\ref{thm:mainK}.]

The UCBs in \eqref{eq:karmucb} imply that $\rad_t(a) \in [\sqrt{\frac{3 \log T}{T}}, \sqrt{3 \log T}]$ for all $a$ and $t$, therefore $\frac{\rmax}{\rmin} \leq \sqrt{T}$.
Then, Propositions~\ref{thm:maingenerale} and \ref{thm:maingeneral} result in the following two inequalities respectively:
\begin{align*}
 \BReg(T, \pi^{\TSUCB}) &\leq 2 \sumt \bE[\rad_t(\At)] + \sqrt{T} + 2\sqrt{\frac{T^{3}}{m}} + T^{-2}, \\
 \BReg(T, \pi^{\TSUCB}) &\leq  2 \sumt \bE[\rad_t(\At)] + \frac{m}{T^2} + T^{-2}.
\end{align*}
Combining these two bounds results in
\begin{align*}
 \BReg(T, \pi^{\TSUCB}) \leq  2 \sumt \bE[\rad_t(\At)] + \sqrt{T} +  T^{-2} +\min\left\{2\sqrt{\frac{T^{3}}{m}}, \frac{m}{T^2}\right\}.
\end{align*}
For any value of $m > 0$, $\min\left\{2\sqrt{\frac{T^{3}}{m}}, \frac{m}{T^2}\right\} \leq 2\sqrt{T}$. Plugging in the known bound for $\sumt \bE[\rad_t(\At)]$ from Theorem~\ref{thm:sqrtbound} finishes the proof of Theorem~\ref{thm:mainK}.\footnote{The statement of Theorem~\ref{thm:mainK} has an additional $+ K$ term since the first $K$ time steps are used to pull each arm once, which we did not include in the proof to simplify exposition.}
\end{myproof}

\begin{myproof}[Proof of Theorem~\ref{thm:mainlinear}.]
The following lemma, whose proof is deferred to Section~\ref{subsub:provelemmas}, allows us to bound $\frac{\rmax}{\rmin}$ by $4\sqrt{2T}$.
\begin{lemma} \label{lem:linradiusbounds}
For the linear bandit, using the UCBs as defined in \eqref{eq:linearucb}, if $||X(a)||_2 =1$ for every $a$, then $\rad_t(a) \in [\subg\sqrt{\frac{d \log T}{T}}, 4\subg\sqrt{2 d \log T}]$ for every $t$ and $a$.
\end{lemma}

Then, using the same steps from the proof of Theorem~\ref{thm:mainK}, we derive
\begin{align}
 \BReg(T, \pi^{\TSUCB}) \leq 2 \sumt \bE[\rad_t(\At)] + 4\sqrt{2T} +  T^{-2} +\min\left\{8 \sqrt{\frac{2T^{3}}{m}}, \frac{m}{T^2}\right\}.
\end{align}

For any $m$, $\min\left\{8 \sqrt{\frac{2T^{3}}{m}}, \frac{m}{T^2}\right\} \leq 8 \sqrt{2T}$. Plugging in the known bound for $\sumt \bE[\rad_t(\At)]$ from Theorem~\ref{thm:sqrtbound_linear} gives us \eqref{eq:mainlineare}, finishing the proof of Theorem~\ref{thm:mainlinear}.
\end{myproof}

\subsection{Deferred Proofs of Lemmas} \label{subsub:provelemmas}
\begin{myproof}[Proof of Lemma~\ref{lem:hoeffding_general}.]
In the linear bandit, this lemma follows directly from Theorem 2 of \cite{abbasi2011improved} (using the parameters $\delta=T^{-3}, \lambda=1$).
In the $K$-armed setting, if $\hmu(n, a)$ is the empirical mean of the first $n$ plays of action $a$, Hoeffding's inequality implies $\Pr(\ft(a) - \hmu(n, a) \geq  \sqrt{\frac{3 \log T}{n}} ) \leq T^{-6}$ for any $n$.
Then, since the number of plays of a particular action is no larger than $T$, we have
\begin{align*}
\Pr(\ft(a) - \hmu_t(a) \geq \sqrt{\frac{3 \log T}{N_t(a)}}) &\leq \Pr(\cup_{n=1}^T \{\ft(a) - \hmu(n, a) \geq \sqrt{\frac{3 \log T}{n}}\}) \leq  T^{-5}.
\end{align*}
Since $|\cA| = K \leq T$ and $\ssA, A_t \in \cA$, the result follows after taking another union bound over actions (which proves a stronger bound of $T^{-4}$).
\end{myproof}

\begin{myproof}[Proof of Lemma~\ref{lem:random}.]
Fix $H_t$ and $\tf_t$. For every action $a$, let $\Delta_a = \tf_t - \hmu_t(a)$, and hence $\Psie_t(a) = \frac{\Delta_a}{\rad_t(a)}$.
Let $\nu$ be a distribution over $\cA_t$.
Then,
\begin{align} \label{eq:lemrand}
\bPsi_t(\nu)  = \frac{\bE_{a \sim \nu}[\Delta_a]}{ \bE_{a \sim \nu}[\rad_t(a)]}.
\end{align}
$\rad_t(a)> 0$ for all $a$, but $\Delta_a$ can be negative.
We claim that the above ratio is minimized when $\tau$ puts all of its mass on one action ---
in particular, the action $\ssa \in \argmin_a \frac{\Delta_a}{\rad_t(a)}$.

For $a \neq \ssa$, let $c_a = \frac{\rad_t(a)}{\rad_t(\ssa)} > 0$.
Then, since $\Psie_t(a) \geq \Psie_t(\ssa)$, we can write
$\Delta_a = c_a\Delta_{\ssa} + \delta_a$ for $\delta_a \geq 0$ for all $a$.
Let $p_{\ssa} = \Pr(a = \ssa)$.
Let $E = \{a \neq \ssa\}$
Substituting into \eqref{eq:lemrand}, we get
\begin{align*}
\bPsi_t(\nu)
&= \frac{\bE[c_a\Delta_{\ssa} + \delta_a]}
        {\bE[c_a \rad_t(\ssa)]} \\
&= \frac{p_{\ssa}\Delta_{\ssa} + \bE[c_a\Delta_{\ssa} + \delta_a|E]\Pr(E)}
        {p_{\ssa} \rad_t(\ssa) +  \bE[c_a \rad_t(\ssa)|E]\Pr(E)} \\
&= \frac{\Delta_{\ssa}\left(p_{\ssa} + \bE[c_a|E]\Pr(E)\right) + \bE[\delta_a|E]\Pr(E)}
        { \rad_t(\ssa)\left( p_{\ssa}+  \bE[c_a|E]\Pr(E)\right)} \\
&= \frac{\Delta_{\ssa}} { \rad_t(\ssa)} +
   \frac{\bE[\delta_a|E]\Pr(E)}{\rad_t(\ssa)\left( p_{\ssa}+  \bE[c_a|E]\Pr(E)\right)} \\
&\geq \frac{\Delta_{\ssa}} { \rad_t(\ssa)} \\
&= \Psi_t(\ssa)
\end{align*}
\end{myproof}

\begin{myproof}[Proof of Lemma~\ref{lem:linradiusbounds}.]
We have
\begin{align*}
\rad_t(a) = \sqrt{\beta_t} || X(a)||_{V_t^{-1}}= \sqrt{\beta_t} ||V_t^{-1/2} X(a)||_2.
\end{align*}
Then, since $||X(a)||_2 = 1$ for all $a$,
\[
\sqrt{\beta_t}\sigma_{\min}(V_t^{-1/2}) \leq \rad_t(a) \leq \sqrt{\beta_t}\sigma_{\max}(V_t^{-1/2}).
\]

First, we lower bound $\sigma_{\min}(V_t^{-1/2})$.
To do this, we can instead upper bound $||V_t||_2$, since $\sigma_{\min}(V_t^{-1/2}) = \sqrt{\sigma_{\min}(V_t^{-1})} = \frac{1}{\sqrt{\sigma_{\max}(V_t)}} = \frac{1}{\sqrt{||V_t||_2}}$.
The triangle inequality gives $||V_t||_2 \leq || I ||_2 + \sum_{s = 1}^t ||X_s X_s^\top||_2$.
Since $X_s X_s^\top$ is a rank-1 matrix, the only non-zero eigenvalue is $||X_s||_2^2 = 1$ with eigenvector $X_s$, since $(X_s X_s^\top) X_s = X_s (X_s^\top X_s)$. Therefore, $||V_t||_2 \leq || I ||_2 + \sum_{s = 1}^t ||X_s||_2^2 \leq 1 + T$, which implies $\sigma_{\min}(V_t^{-1/2})  \geq \frac{1}{\sqrt{T+1}} \geq \frac{1}{\sqrt{2T}}$.
Recall $\sqrt{\beta_t} = \subg \sqrt{d \log(T^2(1+t))} + S \geq \subg \sqrt{d \log T}$, implying $\rad_t(a) \geq \subg \sqrt{\frac{d \log T}{2T}}$.

Next, we upper bound $\sigma_{\max}(V_t^{-1/2}) = \frac{1}{\sqrt{\sigma_{\min}(V_t)}}$ by lower bounding $\sigma_{\min}(V_t)$.
$\sigma_{\min}(V_t) \geq \sigma_{\min}(I) = 1$. Therefore, $\sigma_{\max}(V_t^{-1/2}) \leq 1$.
We can upper bound $\sqrt{\beta_t}$ by $\subg\sqrt{d \log(T^4)} + S \leq 2 \subg\sqrt{4d \log(T)}$, since we assumed $\subg \geq 1$ and $S \leq \sqrt{d}$.
Therefore, we have
\[
\rad_t(a) \leq \sqrt{\beta_t} \leq 4 \subg\sqrt{d \log(T)}.  \;
\]
\end{myproof}